\documentclass[11pt,oneside,english]{amsart}
\usepackage{mathpazo}
\usepackage[scaled=0.92]{helvet}
\usepackage{courier}
\usepackage[T1]{fontenc}
\usepackage[latin9]{inputenc}
\usepackage[letterpaper]{geometry}
\usepackage{algorithm2e}
\usepackage{amstext}
\usepackage{amsthm}
\usepackage{graphicx}
\usepackage{setspace}
\usepackage[authoryear]{natbib}
\makeatletter

\providecommand{\tabularnewline}{\\}

\numberwithin{equation}{section}
\numberwithin{figure}{section}

\theoremstyle{plain}

\numberwithin{equation}{section}
\usepackage{tikz}
\usetikzlibrary{fit,positioning}

\makeatother

\usepackage{babel}
\begin{document}

\title[Partially collapsed sampling for topic models]{ Sparse Partially Collapsed MCMC for Parallel Inference in Topic
Models}

\author{M\r{a}ns Magnusson, Leif Jonsson, Mattias Villani and David Broman}

\thanks{Magnusson: \textit{Link\"{o}ping University} Jonsson: \textit{Ericsson
AB and Link\"{o}ping University} Villani: \textit{Link\"{o}ping University}
Broman: \textit{KTH Royal Institute of Technology}}
\begin{abstract}
Topic models, and more specifically the class of Latent Dirichlet Allocation (LDA), are widely used for probabilistic modeling of text. MCMC sampling from the posterior distribution is typically performed using a collapsed Gibbs sampler. We propose a parallel sparse partially collapsed Gibbs sampler and compare its speed and efficiency to state-of-the-art samplers for topic models on five well-known text corpora of differing sizes and properties. In particular, we propose and compare two different strategies for sampling the parameter block with latent topic indicators. The experiments show that the increase in statistical inefficiency from only partial collapsing is smaller than commonly assumed, and can be more than compensated by the speedup from parallelization and sparsity on larger corpora. We also prove that the partially collapsed samplers scale well with the size of the corpus. The proposed algorithm is fast, efficient, exact, and can be used in more modeling situations than the ordinary collapsed sampler. 
\end{abstract}

\maketitle

\section{Introduction}
\label{sec:intro}

Latent Dirichlet allocation (LDA) \citet{blei2003latent} is an immensely popular\footnote{The original paper has so far been cited roughly 1200 times per year, and the citation rate is sharply increasing after more than ten years since its publication.} way to model text probabilistically. The basic LDA model generates documents as probabilistic mixtures of topics. The observed data is the set of words, or tokens, $\mathbf{w}$, in a given corpus were $w_{i,d}$ is a token at position $i$ in document $d$. Each document $d$ is assigned a vector $\theta_{d}$ which is a probability distribution over $K$ topics. Each topic is a probability distribution $\phi_{k}$ over a vocabulary of word types. Each token at position $i$ in document $d$ is accompanied by a latent topic indicator $z_{i,d}$ generated from $\theta_{d}$, such that $z_{i,d}=k$ means that the token in the $i$th position in document $d$ is generated from $\phi_{k}$. Let $\Theta$ denote the set of all $\theta_{d}$, $\mbox{\textbf{z}}$ all $z_{i,d}$ in all documents, and let $\Phi$ be a $K\times V$ matrix whose $k$th row holds $\phi_{k}$ over a vocabulary of size $V$ of word types $v$. The generative model for LDA can be found in Figure \ref{fig:generative_model} and a summary of model notation in Table \ref{tab:Notation-used.}.

\begin{table}
\begin{centering}
\begin{tabular}{clccl}
\hline 
{\scriptsize{}Symbol} & {\tiny{}Description} &  & {\tiny{}Symbol} & {\tiny{}Description}\tabularnewline
\cline{1-2} \cline{4-5} 
{\scriptsize{}$V$} & {\tiny{}The size of the vocabulary} &  & {\scriptsize{}$\Phi$} & {\tiny{}The matrix with word-topic probabilities : $K\times V$}\tabularnewline
{\scriptsize{}$v$} & {\tiny{}Word type} &  & {\scriptsize{}$\phi_{k}$} & {\tiny{}The word probabilities for topic $k$: $1\times V$}\tabularnewline
{\scriptsize{}$K$} & {\tiny{}The number of topics} &  & {\tiny{}$\beta$} & {\tiny{}The hyperparameter for the prior of $\Phi$: $1 \times 1$}\tabularnewline
{\scriptsize{}$D$} & {\tiny{}The number of documents} &  & {\scriptsize{}$n^{(w)}$} & {\tiny{}The number topic indicators by topic and word type: $K\times V$}\tabularnewline
{\scriptsize{}$N$} & {\tiny{}The total number of tokens} &  & {\tiny{}$\Theta$} & {\tiny{}Document-topic proportions: $D\times K$}\tabularnewline
{\scriptsize{}$N_{d}$} & {\tiny{}The number of tokens in document $d$} &  & {\tiny{}$\theta_{d}$} & {\tiny{}Topic probability for document $d$: $1\times K$}\tabularnewline
{\tiny{}$z_{i,d}$} & {\tiny{}Topic indicator for token $i$ in document $d$} &  & {\tiny{}$\alpha$} & {\tiny{}The hyperparameter for the prior of $\Theta$: $1 \times 1$}\tabularnewline
{\tiny{}$w_{i,d}$} & {\tiny{}Token $i$ in document $d$} &  & {\scriptsize{}$n^{(d)}$} & {\tiny{}The number topic indicators by document and topic: $D\times K$}\tabularnewline
\hline 
\end{tabular}
\par\end{centering}
\caption{LDA model notation. \label{tab:Notation-used.}}
\end{table}

\begin{figure}
\begin{tikzpicture} 
\tikzstyle{main}=[circle, minimum size = 10mm, thick, draw =black!80, node distance = 16mm] 
\tikzstyle{connect}=[-latex, thick] 
\tikzstyle{box}=[rectangle, draw=black!100]
\node[main, draw=none, fill=none] (alpha) {$\alpha$};
\node[main] (theta) [right=of alpha,xshift=-8mm,label=below:$\theta$] { };
\node[main] (z) [right=of theta,label=below:$z$] {};
\node[main, fill = black!10] (w) [right=of z,label=below:$w$] { };
\node[main] (phi) [right=of w,label=below:$\phi$] { };
\node[draw=none,fill=none,right=of phi] (beta) {$\beta$};
\path 
(alpha) edge [connect] (theta)
(theta) edge [connect] (z)         
(z) edge [connect] (w)
(beta) edge [connect] (phi)           
(phi) edge [connect] (w);
\node[rectangle, inner sep=0mm, fit= (z) (w),label=below right:$N$, yshift=1mm, xshift=11.5mm] {};
\node[rectangle, inner sep=-1mm, fit= (phi),label=below right:$K$, xshift=0mm] {};
\node[rectangle, inner sep=0mm, xshift=-25mm, fit= (theta) (z) (w), label=below left:$D$] {};
\node[rectangle, inner sep=4.4mm,draw=black!100, fit= (z) (w)] {};
\node[rectangle, inner sep=5.5mm, draw=black!100, fit = (theta) (z) (w) ] {};
\node[rectangle, inner sep=5mm, draw=black!100, fit = (phi)] {};
\end{tikzpicture}
\begin{enumerate}
\item For each topic $k=1,...,K$

\begin{enumerate}
\item Draw a distribution over words $\phi_{k}\overset{iid}{\sim}\mbox{Dir}_{V}(\beta)$
\end{enumerate}
\item For each observation/document $d=1,...,D$

\begin{enumerate}
\item Draw topic proportions $\theta_{d}|\alpha\overset{iid}{\sim}\mbox{Dir}_{K}(\alpha)$
\item For $i=1,...,N_{d}$ 

\begin{enumerate}
\item Draw topic assignment $z_{i,d}|\theta_{d}\overset{iid}{\sim}\mbox{Categorical}(\theta_{d})$ 
\item Draw token $w_{i,d}|z_{i,d},\phi_{z_{i,d}}\overset{indep}{\sim}\mbox{Categorical}(\phi_{z_{i,d}})$
\end{enumerate}
\end{enumerate}
\end{enumerate}
\caption{The generative LDA model. \label{fig:generative_model}}
\end{figure}

One of the most popular inferential techniques for topic models is Markov Chain Monte Carlo (MCMC) and the collapsed Gibbs sampler introduced by \citet{Griffiths2004}, where both $\Phi$ and $\Theta$ are marginalized out and the elements in $\mathbf{z}$ are sampled by Gibbs sampling. It is a useful building block to use in other more advanced topic models, but it suffers from its sequential nature, which makes the algorithm practically impossible to parallelize in a way that still generates samples from the correct invariant distribution. This sequentiality of the algorithm is a serious problem as textual data are growing at an increasing rate; some recent applications of topic models are counting the number of documents in the billions \citep{yuan2015lightlda}. The computational problem is further aggravated since large corpora typically enable more complex models and a greater number of topics.

The response to these computational challenges has been to use approximations to parallelize the collapsed sampler, such as the popular AD-LDA algorithm by \citet{Newman2009}. AD-LDA samples the latent topic indicators \textbf{z} on different cores in isolation before a synchronization step, thereby ignoring that topic indicators in different documents are dependent after marginalizing out $\Theta$ and $\Phi$. As a result, AD-LDA does not target the true posterior. The total approximation error for the joint posterior is unknown \citep{Ihler2012}, and the only way to check the accuracy in a given application is to compare the inferences to an exact MCMC sampler that is guaranteed to converge to the true posterior distribution.

Instead, we propose a sparse partially collapsed approach to sampling in topic models, resulting in an exact MCMC sampler that will converge to the true posterior. This sampler is achieved by only collapsing over the topic distributions $\theta_{1},...,\theta_{D}$ in each document. The remaining parameters can then be sampled by Gibbs sampling by iterating between the two updates $\mathbf{z}|\Phi$ and $\Phi|\mathbf{z}$, where the topic indicators $z_{i,d}$ are now conditionally independent between documents and the rows of the topic-word matrix $\Phi$ are independent given $\mathbf{z}$. This independence means that the first step can be parallelized with regard to documents, and the second step can be parallelized with regard to topics. Importantly, we also exploit that conditioning on $\Phi$ opens up several elegant ways to take advantage of sparsity and to reduce the time complexity in sampling the $z$'s within a document, as detailed below. Following the literature in the LDA community, we refer to our algorithm as a partially collapsed Gibbs sampler. This should not be confused with the partially collapsed Gibbs samplers of \citet{van2008partially} where different parameters are marginalized in a different step of the Gibbs sampler. All partially collapsed samplers proposed here marginalize out $\Theta$ analytically and sample from the joint posterior of $z$ and $\Phi$ using either Gibbs sampling or Metropolis-Hastings. The hyperparameters in the priors can be sampled in separate updating steps, but we have for simplicity kept them fixed in the analysis.

Partially collapsed and uncollapsed samplers for LDA are noted in \citet{Newman2009}, but quickly dismissed because of lower MCMC efficiency compared to the collapsed sampler. However, the efficiency improvement resulting from collapsing parameters is model specific and must here be weighed against the benefits of parallelization. We show empirically that the efficiency loss from using a partially collapsed Gibbs sampler for LDA compared to a fully collapsed Gibbs sampler is small. This result is consistent across different well-known datasets and for various model settings, a result similar to that found by \citet{NIPS2014_5531} for LDA models using GPU parallelization and by \citet{ishwaran2001gibbs} in the context of Dirichlet process mixtures. Furthermore, we show theoretically, under some mild assumptions, that despite the additional sampling of the $\Phi$ matrix, the complexity of our sampler is still only $O(\sum_{i}^{N}K_{d(i)})$, where $N$ is the total number of tokens in the corpus and $K_{d(i)}$ is the number of existing topics in the document of token $i$. Importantly, the alternative partially collapsed sampler where $\Phi$ is integrated out instead of $\Theta$ (or a fully uncollapsed sampler) will not enjoy the same theoretical scalability with respect to corpus size. We also propose a Metropolis-Hastings based sampler with complexity $O(N)$, similar in spirit to that of light-LDA \citet{yuan2015lightlda}. 

Several extensions and refinements of the partially collapsed sampler are developed to reduce the sampling complexity of the algorithm. For example, we propose a Gibbs sampling version using the Walker-Alias tables proposed in \citet{li2014reducing}, something that is only possible using a partially collapsed sampler. We also note that partial collapsing makes it possible to use more elaborate models on $\Phi$ for which the fully collapsed sampler cannot be applied. As an example, we develop a spike-and-slab prior in the Appendix where we set elements of $\Phi$ to zero using ordinary Gibbs sampling, a type of topic model that previously has been shown to improve topic model performance using variational Bayes inference methods \citep{chien2014bayesian}.

\section{Related work}

The problems of parallelizing topic models have been studied extensively \citep{Ihler2012,Liu2011,Newman2009,Smola2010,Yan2009,Ahmed2012,NIPS2014_5531} together with ways of improving the sampling efficiency of the collapsed sampler \citep{Porteous,Yao2009,li2014reducing,yuan2015lightlda}.

The standard sampling scheme for the topic indicators is the collapsed Gibbs sampler of \citet{Griffiths2004} where the topic indicator for word $i$, $z_{i,d}$, is sampled from

\[
P(z_{i,d}=j|\mathbf{z}_{-i},\mathbf{w})\propto\underbrace{\frac{n_{-i,j,w_{i,d}}^{(w)}+\beta}{n_{-i,j}^{(\cdot)}+V\beta}}_{topic-word}\underbrace{\left(n_{-i,d,j}^{(d)}+\alpha\right)}_{document-topic}\:,
\]

where the scalars $\alpha$ and $\beta$ are prior hyperparameters for $\theta$ and $\phi$: $\theta_{d}\overset{iid}{\sim}\mathit{\mbox{Dir}}(\alpha)$ and $\phi_{k}\overset{iid}{\sim}\mbox{Dir}(\beta)$. $\mathbf{z}_{-i}$ are all other topic indicators in the corpus, $n_{-i,j}^{(\cdot)}$ is the total number of topic indicators in topic $j$, excluding topic indicator $i$. The $n_{-i,j,w_{i,d}}^{(w)}$ is the number of topic indicators for topic $j$ and the word type of token $w_{i,d}$. Similarly, $n_{-i,d,j}^{(d)}$ is the topic indicator count for topic $j$ within the document $d$ that contains token $i$. Both $n_{-i,j,w_{i,d}}^{(w)}$ and $n_{-i,d,j}^{(d)}$ exclude the current topic indicator $z_{i,d}$. This sampler is sequential in nature since each topic indicator is conditionally dependent on \textit{all other} topic indicators in the whole corpus.

The Approximate Distributed LDA (AD-LDA) in \citet{Newman2009} is currently the most common way to parallelize topic models, both between machines (distributed) and using multiple cores with shared memory (multi-core) on one machine. The idea is that each processor or machine works in parallel with a given set of topic counts in the word-topic count matrix $n^{(w)}$. The word-topic matrices at the different processors are synced after each complete update cycle. This approach is an approximation of the collapsed sampler since the word-topic matrix available on each local processor is sampled in isolation from all other processors. The resulting algorithm is not guaranteed to converge to the target posterior distribution, and will in general not do so. However, \citet{Newman2009} find that this approximation works rather well in practice. A bound for the error of the AD-LDA approximation for the sampling of each topic indicators has been derived by \citet{Ihler2012}. They find that the error of sampling each topic indicator increases with the number of topics and decreases with smaller batch sizes per processor and the total data size. They also conclude that the approximation error increases initially during sampling and then levels off to a steady state \citep{Ihler2012}.

The fact that this approach to parallelize the collapsed Gibbs sampler will not converge to the true posterior has motivated our work to develop parallel algorithms for LDA type models that are both exact and fast. Partially collapsed and uncollapsed samplers for LDA have been studied by \citet{NIPS2014_5531} as an alternative approach for GPU parallel topic models where uncollapsed samplers have shown to converge faster than the collapsed sampler.

In addition to parallelizing topic models, there have been a couple of suggestions on how to improve the speed of sampling in topic models. \citet{Yao2009} reduce the iteration steps needed in sampling each token by using that $n^{(w)}$ and $n^{(d)}$ are sparse matrices. They also use the fact that the hyperparameters $\alpha$ and $\beta$ are constant during sampling and that some calculations need to be performed only once per iteration. The idea are developed further by \citet{li2014reducing} who reduce the sampling complexity by combining Walker-Alias sampling (that can be done, amortized, in constant time) together with the sparsity of $n^{(d)}$. This algorithm reduces the complexity of the algorithm to $O(\sum_{i}^{N}K_{d(i)})$, limiting the iterations to the number of topics in each document. But this approach requires Metropolis-Hastings sampling instead of sampling from the full conditional of each topic indicator. \citet{yuan2015lightlda} reduce the complexity further to $O(N)$ per sampling iteration by using a Metropolis-Hastings approach with clever cyclical proposal distributions. All these improvements are for the serial collapsed sampler with AD-LDA needed for parallelization; the resulting algorithms, therefore, all target an approximation to the true posterior distribution, and the total approximation error is unknown. 

Although the examples in this article are focused on the basic LDA model and multi-core parallel inference for larger datasets, our ideas are easily extended to a broader class of models. First of all, these ideas can easily be used in other more elaborate topic models such as \citet{rosen2010learning}.  Second, it can be used in predicting topic distributions in out-of-corpus documents for predictions in supervised topic models (see \citet{zhu2013gibbs} for an example). Third, it can be used to evaluate topic models \citep{wallach2009evaluation}. The same ideas can also be exploited in other models based on the multinomial-Dirichlet conjugacy properties outside the class of topic models such as Gibbs samplers for part-of-speech tagging \citep{gao2008comparison}.

\section{Partially Collapsed sampling for topic models}

\subsection{The basic partially collapsed Gibbs sampler}

The basic partially collapsed sampler simulates from the joint posterior of $\mathbf{z}$ and $\Phi$ by iteratively sampling from the conditional posterior $p(\mathbf{z}|\Phi)$ followed by sampling from $p(\Phi|\mathbf{z})$. Note that the topic proportions $\theta_{1},...,\theta_{D}$ have been integrated out in both updating steps and that both conditional posteriors can be obtained in analytical form due to conjugacy. Theadvantage of only collapsing over the $\theta$'s is that the update from $p(\mathbf{z}|\Phi)$ can be parallelized over documents (since they are conditionally independent under this model). In a similar way, the update from $p(\Phi|\mathbf{z})$ can be parallelized over topics (the rows of $\Phi$ are conditionally independent). These properties gives the following basic sampler where we first sample the topic indicators for each document in parallel as

\begin{equation}
p(z_{i,d}=j|\mathbf{z}_{-i,d},\Phi,w_{i,d})\propto\phi_{j,w_{i,d}}\cdot\left(n_{-i,d,j}^{(d)}+\alpha\right)\label{eq:sample_z_pclda}
\end{equation}
where $\mathbf{z}_{-i,d}$ are all topic indicators in document $d$
excluding topic indicator $i$, and then sample the rows of $\Phi$ in parallel as 
\begin{equation}
p(\phi_{k}|\mathbf{z})\sim\mbox{Dir}(\mathbf{n}_{k}^{(w)}+\beta)\,.\label{eq:sample_phi_pclda}
\end{equation}

In the following subsections, we propose a number of improvements of the basic partially collapsed Gibbs sampler to reduce the complexity of the algorithm and to speed up computations. We will present the samplers for a symmetric hyperparameter $\alpha$; extending it to an asymmetric prior with different $\alpha$:s for different topics is straight forward.

\subsection{The sparse partially collapsed Gibbs sampler}

The sampling of $p(\mathbf{z}|\Phi)$ in the basic partially collapsed Gibbs sampler is of complexity $O(K)$ per topic indicator, making the sampling time linear in the number of topics. We propose a sparse partially collapsed Gibbs sampler (PC-LDA) with several improvements of the sampling algorithm.

The Alias-LDA method in \citet{li2014reducing} exploits the sparsity that is created by the topic model when each document only contains a small subset of different topics. This usage of topic sparsity in documents can be extended to the partially collapsed sampler by decomposing

\begin{eqnarray*}
p(z_{i,d}=j|\mathbf{z}_{-i,d},\Phi,w_{i,d}) & \propto & \phi_{j,w_{i,d}}\cdot\left(\alpha+n_{-i,d,j}^{(d)}\right)\\
 & = & \phi_{j,w_{i,d}}\cdot\alpha+\phi_{j,w_{i,d}}\cdot n_{-i,d,j}^{(d_{i})}\:.
\end{eqnarray*}

To sample a topic indicator for a given token we first need to calculate the normalizing constant 

\[
q_{w_{i,d}}(z)=\sum_{k=1}^{K}\phi_{k,w_{i,d}}\cdot\left(\alpha+n_{-i,d,k}^{(d)}\right)=\sigma_{a,w_{i,d}}+\sigma_{b,w_{i,d}}\:,
\]

where $\sigma_{a,w_{i,d}}=\sum_{k=1}^{K}\phi_{k,w_{i,d}}\alpha$ and $\sigma_{b,w_{i,d}}=\sum_{k=1}^{K}\phi_{k,w_{i,d}}n_{-i,d,k}^{(d)}$. 

The importance of this decomposition is two-fold. First, following sparse-LDA \citep{Yao2009}, we can use the sparsity of the topic counts $n^{(d)}$ within a document to calculate $\sigma_{b,w_{i,d}}$ by only iterating over the non-zero topic counts. This iteration reduces the complexity of sampling one topic indicator from $O(K)$ to $O(K_{d})$, where $K_{d}$ is the number of non-zero topics in a given document.
Second, following Alias-LDA by \citet{li2014reducing}, we can exploit that $\sigma_{a,w_{i,d}}$ is constant over the sampling of topic indicators. Therefore we only need to compute $\sigma_{a,w_{i,d}}$ once for each sampling iteration and once for each word type $v$ resulting in an amortized $O(1)$ algorithm (i.e., an algorithm which is $O(1)$ for each $z_{i,d}$ after an initial cost common to all $z_{i,d}$ in a corpus). More specifically, drawing a single $z_{i,d}$ is performed as follows.

First calculate $\sigma_{b,w_{i,d}}$ and the cumulative sum, $\mathbf{s}_{b}=\sum_{k\in n_{-i,k}^{(d_{i})}>0}\phi_{k,w_{i,d}}n_{-i,d,k}^{(d)}$ over non-zero topics in the document. Draw a $U\sim\mathcal{U}(0,\sigma_{a,w_{i,d}}+\sigma_{b,w_{i,d}})$. If $U\leq\sigma_{a,w_{i,d}}$, we use the Walker-Alias method presented in \citet{li2014reducing} to sample a topic indicator with $O(1)$ amortized. The Walker-Alias table method is a method to generate samples from an arbitrary categorical distribution efficiently. The method first constructs an Alias table in $O(K)$ time that it is used to draw a sample in $O(1)$ time \citet{walker1977efficient}. Algorithmic details on how to construct an Alias table and draw a sample from it can be found in Algorithm \ref{alg:ConstructAliasTable()} and \ref{alg:LookUpAliasTable()} in the Appendix. If $U>\sigma_{a,w_{i,d}}$, we choose a topic indicator using binary search over $\mathbf{s}_{b}$ with complexity $O(\log(K_{d}))$ \citet{Xiao2010}. Overall, sampling a topic indicator with PC-LDA, therefore, has complexity $O(K_{d})$ amortized. The full algorithm is described in Algorithm \ref{alg:Sparse-Partially-Collapsed} and \ref{alg:SampleTopicIndicatorsSparse()} in the Appendix. 

Conditioning on $\Phi$ gives us a couple of advantages compared to the original Walker-Alias method for the collapsed sampler. First, the Walker-Alias method can be used in a Gibbs sampler for each topic indicator, unlike the approach of \citet{yuan2015lightlda} that uses a proposal in a Metropolis-Hasting within Gibbs algorithm. Direct simulation from a full conditional is generally more efficient than sampling from the full conditional posterior using a Metropolis-Hastings update (except in the very rare case where the proposal is explicitly set up to generate negatively autocorrelated draws, which is not the case here). Second, as a by-product of calculating the Walker-Alias tables, we also calculate the normalizing constants $\sigma_{a,w_{i,d}}$ for all word types that can be stored and reused in sampling $\mathbf{z}$. Note that building the Alias table can also easily be parallelized by word type.

To sample the Dirichlet distribution (as a normalized sum of gamma distributed variables), we use the method of \citet{marsaglia2000simple} to generate gamma variables efficiently. With this sampler we can take advantage of the sparsity in the $n^{(w)}$ count matrix and increase the speed further by storing previous calculations when sampling $\phi$ for $n^{(w)}=0$.

Another advantage of the PC-LDA approach in a multi-core setting is that since the topic indicators of different documents, $\mathbf{z}_{d}$, are conditionally independent given $\Phi$ it allows us to rearrange document sampling between cores freely during the sampling of $p(\mathbf{z}|\Phi)$. By using a job stealing approach where workers that have finished sampling "their" documents can "steal" jobs from other cores we can balance the workload between workers during sampling \citep{Lea2000}. This approach can probably be improved further, but it shows another straight-forward benefit from conditioning on $\Phi$.

\subsection{The light partially collapsed conditional Metropolis-Hastings sampler}

\citet{yuan2015lightlda} propose an alternative approach to sample topic models with larger $K$ using Metropolis-Hastings (MH) sampling. They use a cyclical proposal distribution, alternating between a word proposal and document proposal to reduce sampling complexity. This approach showed great improvements in the distributed situation and can be straightforwardly extended to a partially collapsed sampler as follows. 

The word-proposal distribution of the proposed topic indicator $z_{i,d}^{*}$ is 

\[
p_{w}(z_{i,d}^{*}=j|\Phi,w_{i,d})\propto\phi_{j,w_{i,d}}\:.
\]

This proposal can be sampled using the Walker-Alias method with complexity $O(1)$ given a constructed Alias-table based on the word types in $\Phi$ in a similar fashion as the sparse sampler in the previous subsection. The acceptance probability of the proposed topic indicator $z_{i,d}^{*}$ is given by 

\begin{eqnarray*}
\pi_{w,i} & = & \min\left\{ 1,\frac{p(z_{i,d}^{*}|\Phi,\mathbf{z}_{-i,d},w_{i,d})p_{w}(z_{i,d}|\Phi,w_{i,d})}{p(z_{i,d}|\Phi,\mathbf{z}_{-i,d},w_{i,d})p_{w}(z_{i,d}^{*}|\Phi,w_{i,d})}\right\} =\min\left\{ 1,\frac{\alpha+n_{-i,d,z_{i,d}^{*}}^{(d)}}{\alpha+n_{-i,d,z_{i,d}}^{(d)}}\right\} ,
\end{eqnarray*}

where $z_{i,d}$ is the current draw, $p(\cdot)$ is the full conditional posterior, $p_{w}(\cdot)$ is the word-proposal distribution and $n_{-i,d,z_{i,d}^{*}}^{(d_{i})}$ is the number of topic indicators in document $d$ and for topic $z_{i,d}^{*}$ but with the $i$th topic indicator excluded. If the proposed topic is more common in the document than the current topic indicator it will be accepted with probability 1. Otherwise, the acceptance probability will be roughly proportional to the ratio $n_{d,z_{i}^{*}}/n_{d,z_{i}}$.

The second proposal in the sampler is the doc-proposal distribution. This is exactly the same proposal distribution as in \citet{yuan2015lightlda} and is given by

\[
p_{d}(z_{i,d}^{*}=j|\mathbf{z}_{d},w_{i,d})\propto n_{d,j}^{(d)}+\alpha\:.
\]

This proposal is sampled using a two-phase approach. First draw $U\sim\mathcal{U}(0,\sum_{k}^{K}\alpha+n_{d,k})$. If $U\leq\alpha K$ we sample a topic indicator with $O(1)$ from $p(z_{i,d}^{*})\propto\alpha$. If $U>\alpha K$ we propose $z_{i,d}^{*}$ proportional to the distribution of topic indicators within the document by simply sample an existing topic indicator as $U(1,N_{d})$ where $N_{d}$ is the number of tokens in document $d$. In this way, we will draw with complexity $O(1)$ from the proposal distribution $p(z_{i,d}^{*})\propto n_{d,z_{i,d}^{*}}$ without any need to create an Alias table. The acceptance probability is given by 
\[
\pi_{d,i}=\min\left\{ 1,\frac{\phi_{z_{i,d}^{*},w_{i,d}}(\alpha+n_{-i,d,z_{i,d}^{*}}^{(d)})(\alpha+n_{d,z_{i,d}}^{(d)})}{\phi_{z_{i,d},w_{i,d}}(\alpha+n_{-i,d,z_{i,d}}^{(d)})(\alpha+n_{d,z_{i,d}^{*}}^{(d)})}\right\} .
\]

To simplify further we can do a slight change in the document proposal and instead propose with

\[
\tilde{p}_{d}(z_{i,d}^{*}|\mathbf{z}_{-i,d},w_{i,d})\propto n_{-i,d,z_{i,d}^{*}}^{(d)}+\alpha_{z_{i,d}^{*}}\:.
\]

Using this proposal distribution we will end up with the simplified acceptance probability 
\[
\tilde{\pi}_{d,i}=\min\left\{ 1,\frac{\phi_{z_{i,d}^{*},w_{i,d}}}{\phi_{z_{i,d},w_{i,d}}}\right\} .
\]

This simplification can also be done in the original light-LDA document proposal acceptance step. In our experiments, we use $\pi_{d,i}$ to enable a fair comparison with the original light-LDA sampler. 

These two proposals are then combined to a cyclical Metropolis-Hastings proposal where the two proposals are used for each topic indicator $z_{i,d}$ in each Gibbs iteration. The benefit of this sampler is that the sampling complexity is reduced compared to the sparse approaches. But the downside is the inefficiency of the sampling and the fact that for each token it can be necessary to draw as many as four uniform variables, a relatively costly (but constant) operation. A complete description of the algorithm can be found in Algorithm \ref{alg:Light-Partially-Collapsed} and \ref{alg:SampleTopicIndicatorsLight()} in Appendix \ref{subsec:Algorithms}.

\subsection{Time complexity of the sampler}

The original collapsed sampler of \citet{Griffiths2004} has sampling complexity $O(N\cdot K)$ per iteration. By taking advantage of sparsity, the complexity of sparse-LDA in \citet{Yao2009} is reduced to $O(\sum_{i}^{N}\max_{i}(K_{d(i)},K_{w(i)}))$ where $K_{d(i)}$ is the number of existing topics in the document of token $i$ and $K_{w(i)}$ is the number of existing topics in the word type $w$. This complexity will reduce to $O(N\cdot K)$ and as $N\rightarrow\infty$ \citep{li2014reducing}. The light sampler proposed by \citet{yuan2015lightlda} has the good property of being of complexity $O(N)$ since sampling each topic indicator can be done
in constant time.

Sampling the topic indicators for the sparse PC-LDA sampler has complexity $O(\sum_{i}^{N}K_{d(i)})$ and the light-PC-LDA sampler $O(N)$, but unlike the fully collapsed sampler, we also need to sample $\Phi$. It is therefore of interest to study how the $\Phi$ matrix (of size $K\times V$) grows with the number of tokens ($N$), to determine the overall complexity of PC-LDA. In most languages, the number of word types follows quite closely to Heaps' law \citet{heaps1978irc}, which models the relationship between word types and tokens as $V(N)=\xi N^{\varphi}$ where $0<\varphi<1$. Typical values of $\varphi$ lies in the range of 0.4 to 0.6 and $\xi$ varies between 5 and 50 depending on the corpus \citep{araujo1997large}.

The number of topics, $K$, in large corpora is often modeled by a Dirichlet process mixture where the expected number of topics are $E(K(N))=\gamma\log\left(1+\frac{N}{\gamma}\right)$, where $\gamma$ is the prior precision of the Dirichlet process \citep{teh2010dirichlet}.

\smallskip{}

\paragraph*{Proposition 1. Complexity of the sparse partially collapsed Gibbs
sampler\label{Prop1}}

\textit{Assuming a vocabulary size following Heaps' law $V=\xi N^{\varphi}$ with $\xi>0,0<\varphi<1$ and the number of topics following the mean of a Dirichlet process mixture $K=\gamma\log\left(1+\frac{N}{\gamma}\right)$ with $\gamma>0$, the complexity of the PC-LDA sampler is} 

\[
O\left(\sum_{i=1}^{N}K_{d(i)}\right).
\]

\begin{proof} See Appendix \ref{subsec:Proofs}.\end{proof}

\paragraph*{Proposition 2. Complexity of the light partially collapsed conditional
Metropolis-Hastings sampler\label{Prop2}}

\textit{Assuming a vocabulary size following Heaps' law $V=\xi N^{\varphi}$ with $\xi>0,0<\varphi<1$ and the number of topics following the mean of a Dirichlet process mixture $K=\gamma\log\left(1+\frac{N}{\gamma}\right)$ with $\gamma>0$, the complexity for the light-PC-LDA sampler is}

\[
O\left(N\right).
\]

\begin{proof} See Appendix \ref{subsec:Proofs}.\end{proof}

Propositions \ref{Prop1} and \ref{Prop2} show that the proposed partially collapsed samplers have an equivalent computational complexity as the state-of-the-art fully collapsed samplers. The sampling of $\Phi$ is dominated by the sampling of $\mathbf{z}$ when $N\rightarrow\infty$.

These result also shed light on the importance of integrating out $\Theta$. The $\Theta$ parameters will grow much faster than $\Phi$ as $N\rightarrow\infty$. This property of $\Theta$ makes the partially collapsed sampler, where we integrate out $\Theta$, the only viable option for larger corpora if we want a sampler with minimal computational complexity.

\section{Experiments\label{sec:Experiments}}

In the following sections, we study the characteristics of the PC-LDA samplers. We compare our algorithm with the sparse-LDA from \citet{Yao2009} parallelized using AD-LDA in Mallet 2.0.7 (called AD-LDA in the experiments below) \citep{mccallum2002mallet}. Note that AD-LDA reduces to an exact sparse-LDA collapsed sampler when we are only using one core. The code for PC-LDA has been released as open source as a plug-in to the Mallet framework.\footnote{\texttt{https://github.com/lejon/PartiallyCollapsedLDA}.}

We use the same corpora as is used by \citet{Newman2009} to evaluate our PC-LDA sampler. We also use the New York Times corpus and a Wikipedia corpus to be able to compare with the results in \citet{hoffman2013stochastic}. Following common practice, we remove the rarest word types in the corpus. We choose a rare word limit of 10 for the smaller corpora. For the larger corpora, we instead follow \citet{hoffman2013stochastic} and use TF-IDF to choose the most relevant vocabulary, using 50 000 terms for the PubMed corpus, 7 700 for the Wikipedia corpus, and 8 000 for the New York Times corpus.

\begin{table}
\caption{Summary statistics of training corpora.}

\noindent\begin{minipage}[t]{1\columnwidth}%
\label{tab:Datasets} 
\begin{center}
\begin{tabular}{lrrr}
\textbf{DATASET}  & \textbf{N}  & \textbf{D}  & \textbf{V} \tabularnewline
\hline &  &  & \tabularnewline
NIPS\footnote{http://archive.ics.uci.edu/ml/datasets/Bag+of+Words }  & \textasciitilde{} 1.9 m & 1 499 & 11 547\tabularnewline
Enron$^{a}$ & \textasciitilde{} 6.4 m & 39 860 & 27 791\tabularnewline
Wikipedia \footnote{The tokenized version has been used. \citet{reese2010word}

http://www.cs.upc.edu/\textasciitilde{}nlp/wikicorpus/tagged.en.tgz} & \textasciitilde{} 157 m & \textasciitilde{} 1.36 m & 7 700\tabularnewline
New York Times \footnote{\citet{sandhaus2008new} https://catalog.ldc.upenn.edu/LDC2008T19 } & \textasciitilde{} 400 m & \textasciitilde{} 1.83 m & 8 000\tabularnewline
PubMed$^{a}$ & \textasciitilde{} 761 m & \textasciitilde{} 8.2 m & 50 000\tabularnewline
\end{tabular}
\par\end{center}%
\end{minipage}
\end{table}

\begin{figure}[h]
\vspace{0.3in}
 
\begin{centering}
\includegraphics[scale=0.15]{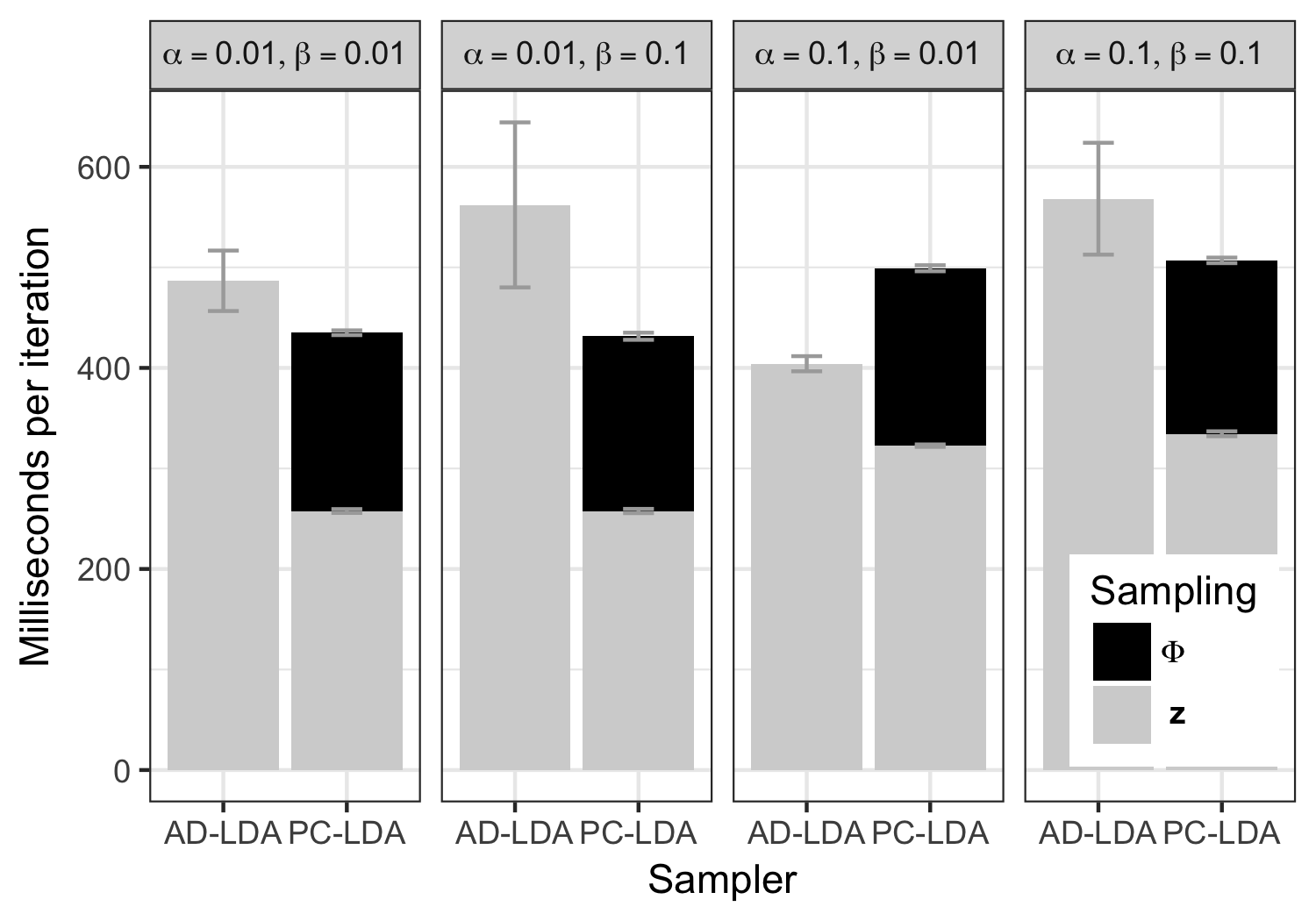} 
\par\end{centering}
\vspace{0.3in}
 \caption{Average time per iteration (incl. standard errors) for Sparse AD-LDA
and for PC-LDA using the Enron corpus and 100 topics.\label{fig:prior_speed}}
\end{figure}

The choice of hyperparameters influences the sparsity of $n^{(w)}$ and $n^{(d)}$, and hence also the relative speed of the studied samplers: sparse AD-LDA is mainly fast for sparse $n^{(w)}$ while PC-LDA benefits from the sparsity of $n^{(d)}$. Since $\alpha$ influences the sparsity of $n^{(d)}$ while $\beta$ influences the sparsity of $n^{(w)}$ we, therefore, ran experiments comparing the computing time per iteration. We ran the samplers for different combinations of $\alpha$ (0.1 and 0.01) and $\beta$ (0.1 and 0.01) for the Enron corpus with $K=100$ topics. These experiments were performed with six different initializations and the sampling time at the $1000$th iteration for all six seeds. Figure \ref{fig:prior_speed} shows that the PC-LDA sampler is only slower than Sparse AD-LDA when $\alpha=0.1$ and $\beta=0.01$. To ensure that our results are conservative, we set $\alpha=0.1$ and $\beta=0.01$ in all experiments to put PC-LDA in the least favorable situation.

Finding suitable metrics for comparing sampler for topics models is a challenge for a number of reasons. First, our samplers (sparse PC-LDA and light PC-LDA) are proper MCMC methods known to converge to the target posterior. This is not true for the samplers that we compare to (AD-LDA and Light LDA using AD-LDA for parallelization) that at best converge to a reasonable approximation of the posterior. However, there is currently no theory to back up this claim. It is therefore not possible to compare samplers using the usual metrics from the MCMC literature (e.g., integrated autocorrelation time), except when using a single processor (in which case AD-LDA and Light-LDA are proper MCMC methods converging to the target posterior). Second, topic models are highly complex models with millions or even billions of latent discrete variables learned jointly with other high-dimensional continuous parameters. Like in any mixture model, the posterior is expected to have many local minor modes (even without considering the so-called label switching problem), and it is practically impossible to explore the full parameter space in any reasonable amount of time. The goal for any posterior sampling method in such models is therefore 

\begin{enumerate}
\item to quickly locate the regions of dominant posterior mass and 
\item to efficiently explore those major modes in proportions to their posterior
density. 
\end{enumerate}

The first aim has not been studied much in the theoretical MCMC literature, except in explicit MCMC-based optimization algorithms such as simulated annealing where the second goal is not reached. One exception is \citet{maciuca2006first} who study the mean first hitting time of the independence Metropolis-Hastings algorithm (i.e., the expected time to reach a given point in the parameter space). Only studying the first aim of the posterior sampling method is though limited in evaluating Markov Chains. Hence, we will, therefore, analyze both the algorithm's ability to find the dominant modes quickly and its mixing properties via the integrated autocorrelation time. The integrated autocorrelation time does not depend on the number of parallel processors, and it is, therefore, sufficient to compute it for the single processor case. As noted above, it also does not make sense to calculate it for methods using AD-LDA in the multi-processor setting.

Following the evaluation of topic models in the machine learning literature (see also \citet{villani2009regression} for similar evaluations for mixture-of-experts models), we evaluate the samplers and how well they reach the regions of high posterior density as well as the mixing properties using the integrated autocorrelation time. We are using the log joint posterior of the topic indicators ($\text{\textbf{z}}$) with $\Phi$ and $\Theta$ marginalized out; we refer to this quantity as the log marginalized posterior. Focusing only on the topic indicators makes the evaluation comparable across all algorithms. Since the behavior of the chain depends on the initialization state, we have used the same seed to initialize the different samplers to the exact same starting state (concerning $\mathbf{z}$).

The experiments are conducted using an HP Cluster Platform with DL170h G6 compute nodes with 4-core Intel Xeon E5520 processors at 2.2GHz (for the 8-core experiments) or 8-core Intel Xeon E5-2660 "Sandy Bridge" processors at 2.2GHz (speed experiments). All experiments use two sockets with 24 or 32 GB memory nodes, except for the parallelism experiment where we use an 8-socket 64-core machine with 1024 GB memory.

\subsection{Efficiency loss from only partially collapsing\label{subsec:Inefficiency}}

\citet{liu1994collapsed} proves that collapsing out parameters improves the mixing rate of the MCMC chain for the remaining parameters. Contrary to often held beliefs (see, e.g., \citet{Newman2009}) Theorem 1 in \citet{liu1994collapsed} is not applicable to LDA when $\Phi$ (or $\Theta$) is integrated out. This fact has recently been pointed out by \citet{TereninPreparation} who give a simple counterexample to demonstrate this point. It is, therefore, an open question whether the mixing rate of a partially collapsed Gibbs sampler is worse than a fully collapsed sampler in the LDA context, and if so, by how much. We will here investigate this empirically on two well-known corpora where we compare the inefficiency factor (integrated autocorrelation time) of the fully collapsed and the PC-LDA sampler in a single core setting. 

Each experiment starts with a given random seed and runs for 10 000 iterations using the collapsed Gibbs sampler (the gold standard) to justify that we have reached the posterior region of interest to explore using a visual inspection of the traceplot for the log marginalized posterior. The topic indicators $\mathbf{z}$ in the last iteration is then used as initialization point for both a collapsed sampler and the PC-LDA sampler. We subsequently perform two sub-runs with the collapsed sampler and two with the PC-LDA sampler per experimental setup.

The parameters $\Theta$ and $\Phi$ are subsequently sampled for each of the 2 000 $\mathbf{z}$-draws. For the collapsed sampler $p(\Theta,\Phi\vert\mathbf{z})$ are sampled while we sample $p(\Theta\vert\mathbf{z})$ for the PC-LDA sampler (we already have samples of $\Phi$). We calculate the inefficiency for the 1 000 largest mean values of $\phi_{k}$ for each topic (the so-called top words) and all elements in $\theta_{d}$ for 1000 randomly chosen documents. This means that the inefficiency estimates are based on $1000\cdot K$ parameters for $\Theta$ (Table \ref{tab:Inefficiency-of-Theta-Table}) and $1000\cdot K$ for $\Phi$ (Table \ref{tab:Inefficiency-of-Phi-Table}).
To estimate the inefficiency factor (IF) for each parameter, we compute $\text{IF}=L/ESS$ where $L$ is the length of the Markov Chain and $ESS$ is effective sample size computed with the \texttt{coda} package \citep{Plummer:2006aa} in R. Up to a certain lag, we find this package to be much more precise than the estimator based on sample autocorrelations.

\begin{table}[h]
\caption{Mean inefficiency factors, IF, (standard deviation in parentheses)
of $\Theta$.}
\label{tab:Inefficiency-of-Theta-Table} 
\centering{}%
\begin{tabular}{lrrrr}
\textbf{DATA}  & \textbf{K}  & \textbf{Collapsed}  & \textbf{PC-LDA}  & \textbf{IF ratio}\tabularnewline
\hline &  &  &  & \tabularnewline
Enron  & 20  & 3.31 (4.8)  & 3.54 (6.1)  & 1.07\tabularnewline
Enron  & 100  & 2.21 (5.0)  & 2.29 (5.3)  & 1.04\tabularnewline
NIPS  & 20  & 10.82 (32.0)  & 12.54 (47.2)  & 1.16\tabularnewline
NIPS  & 100  & 6.64 (14.1)  & 7.45 (16.0)  & 1.12\tabularnewline
\end{tabular}
\end{table}

\begin{table}[h]
\caption{Mean inefficiency factors, IF, (standard deviation in parentheses)
of $\Phi$.}
\label{tab:Inefficiency-of-Phi-Table} 
\centering{}%
\begin{tabular}{lrrrr}
\textbf{DATA}  & \textbf{K}  & \textbf{Collapsed}  & \textbf{PC-LDA}  & \textbf{IF ratio}\tabularnewline
\hline &  &  &  & \tabularnewline
Enron  & 20  & 5.00 (19.9)  & 5.03 (14.7)  & 1.01\tabularnewline
Enron  & 100  & 17.90 (49.2)  & 22.46 (58.0)  & 1.26\tabularnewline
NIPS  & 20  & 28.20 (73.5)  & 31.47 (81.1)  & 1.12\tabularnewline
NIPS  & 100  & 16.20 (43.1)  & 23.85 (55.6)  & 1.48\tabularnewline
\end{tabular}
\end{table}

The results of the first experiment can be seen in Table \ref{tab:Inefficiency-of-Theta-Table} and \ref{tab:Inefficiency-of-Phi-Table}; the other experiments gave very similar inefficiencies and are not reported. We conclude that the increase in inefficiency of the chains from not collapsing out $\Phi$ is small. The largest value, 1.48, can be found in the NIPS dataset. In Figure \ref{fig:loglik_nips20}, we can see the effect this has on the speed of the chain to reach the region of high posterior density. Note that while a partially collapsed sampler has nearly the same mixing properties as the collapsed sampler, it can be parallelized to run substantially faster in a multi-core setting; this is demonstrated in Section \ref{subsec:Speedup-and-parallelism}.

\subsection{Posterior error using the AD-LDA approximation\label{subsec:Full-posterior-error-ADLDA}}

AD-LDA is the most popular way of parallelizing LDA. It is known that the approximation will influence the sampling of each topic indicator \citep{Ihler2012}, but we have not found any studies of the effect on the joint posterior distribution. To explore this, we start the sampler with the same initial state with respect to the topic indicators $\mathbf{z}$ and then run the sampler with different numbers of cores/partitions to see the effect on the joint posterior distribution of the topic indicators $p(\mathbf{z}\vert\mathbf{w})$.

\begin{figure}[h]
\vspace{0.3in}
 
\begin{centering}
\includegraphics[scale=0.14]{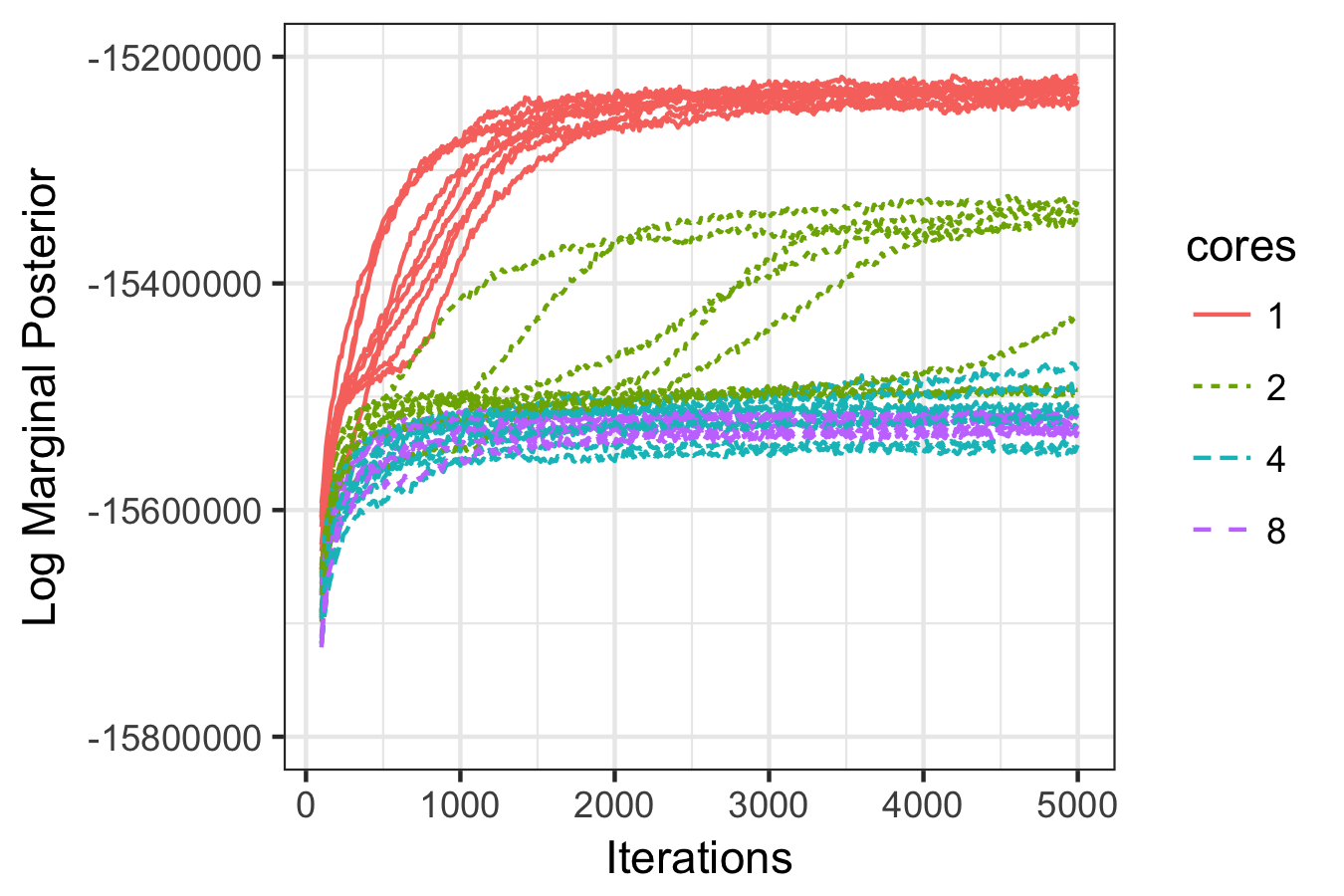}
\includegraphics[scale=0.14]{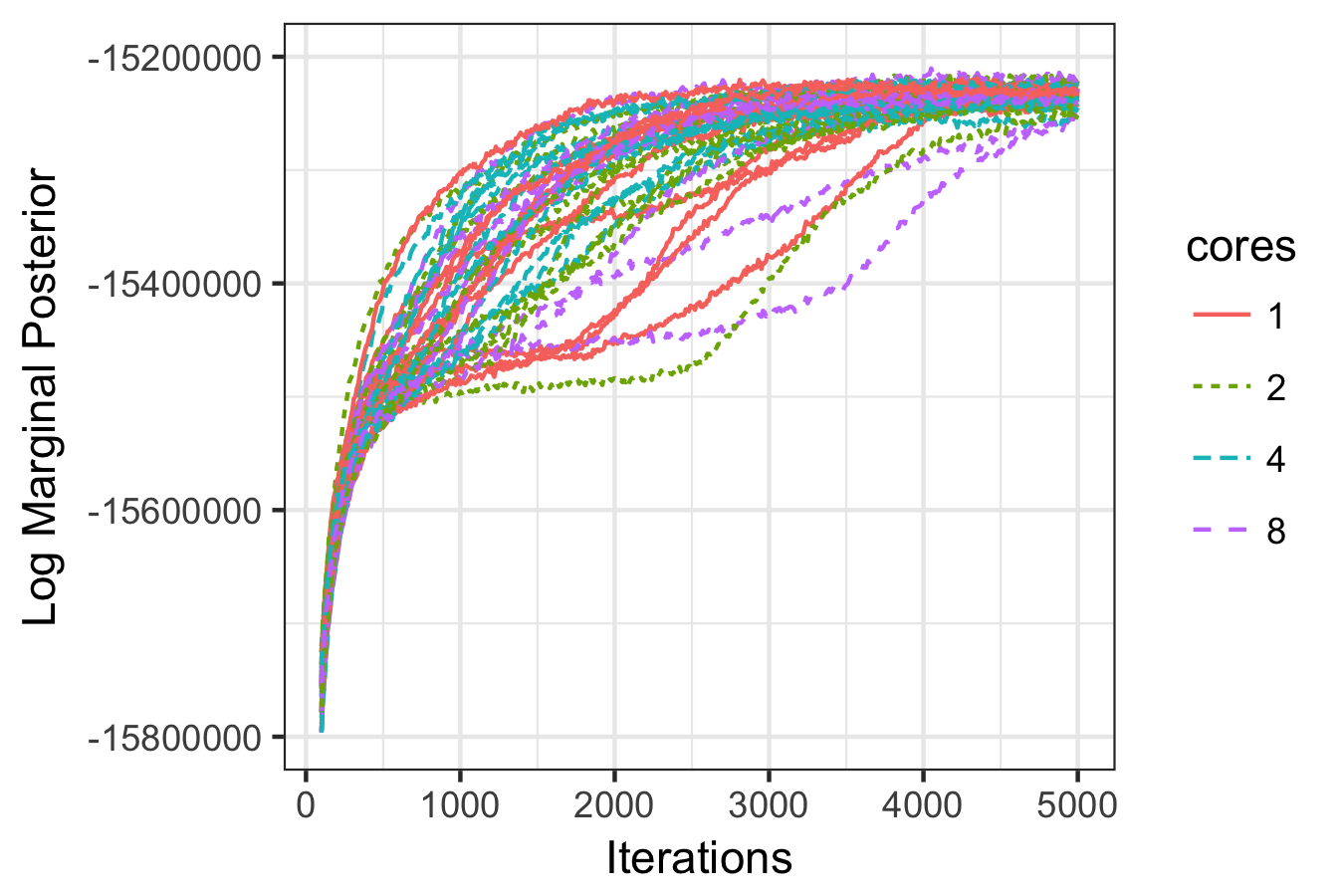} 
\par\end{centering}
\begin{centering}
\includegraphics[scale=0.14]{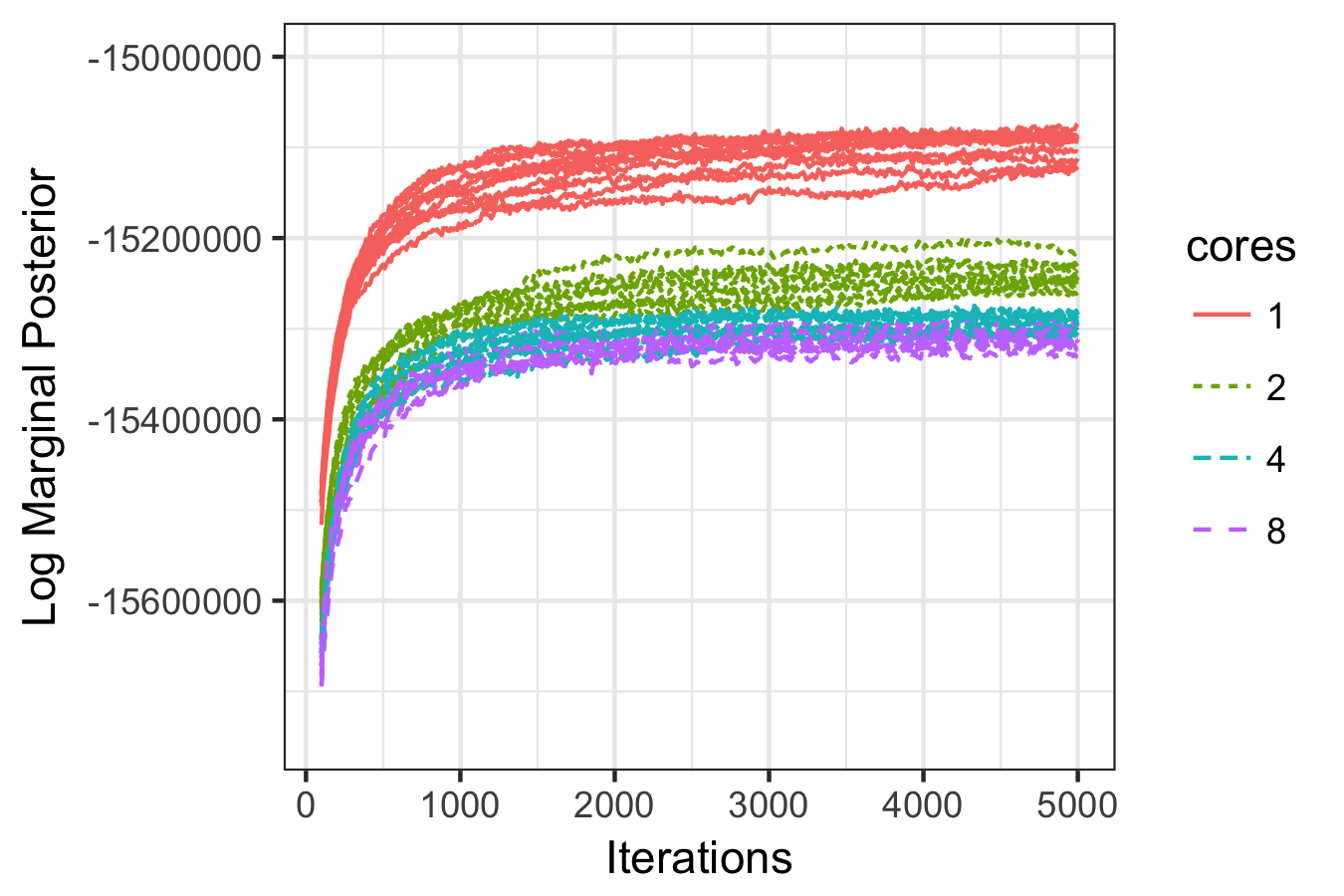}
\includegraphics[scale=0.14]{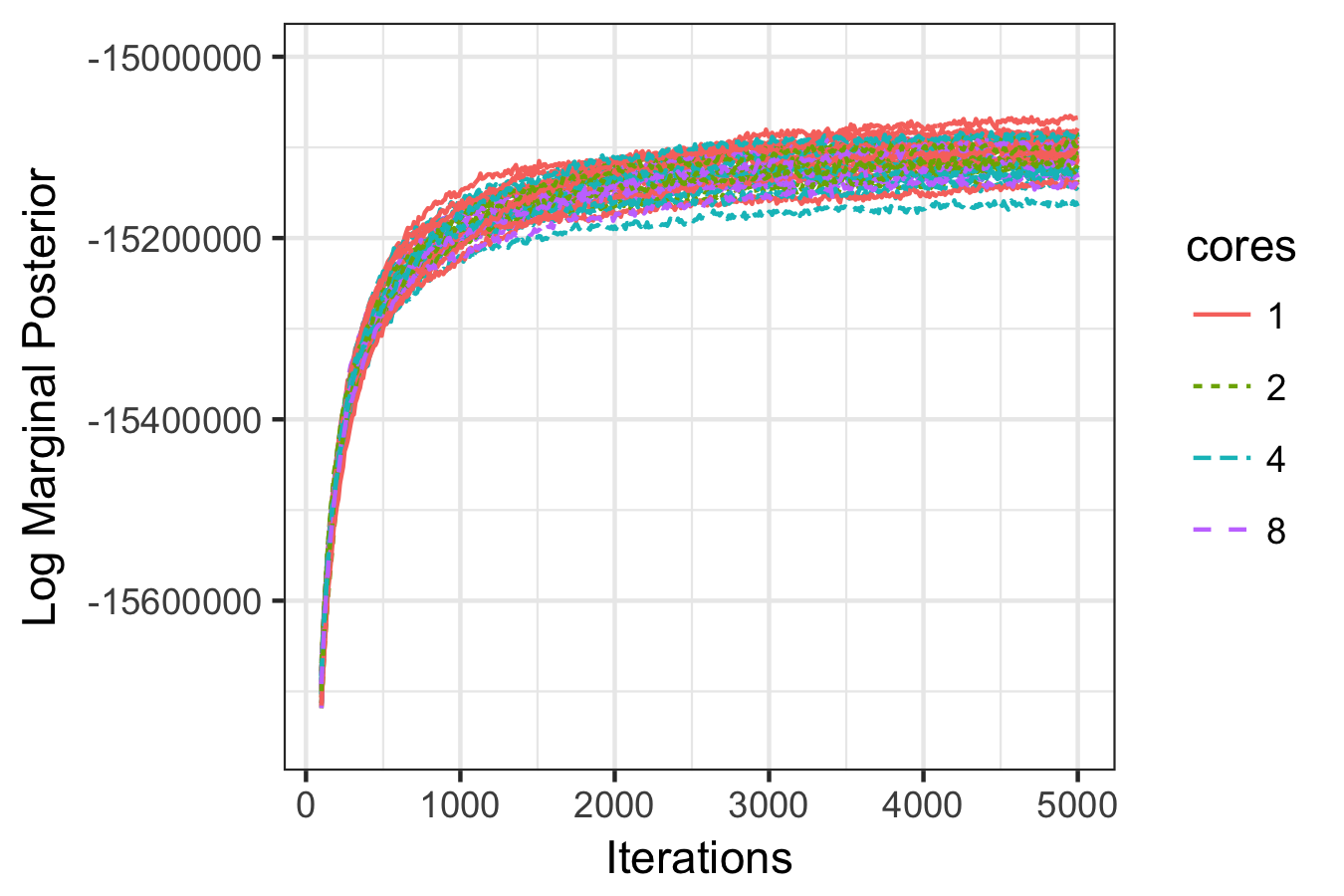}
\par\end{centering}
\vspace{0.3in}
 \caption{Log marginalized posterior for the NIPS dataset with $K=20$ (upper)
and $K=100$ (lower) for AD-LDA (left) and PC-LDA (right).}
\label{fig:loglik_nips20} 
\end{figure}

\begin{figure}[h]
\vspace{0.3in}
 
\begin{centering}
\includegraphics[scale=0.14]{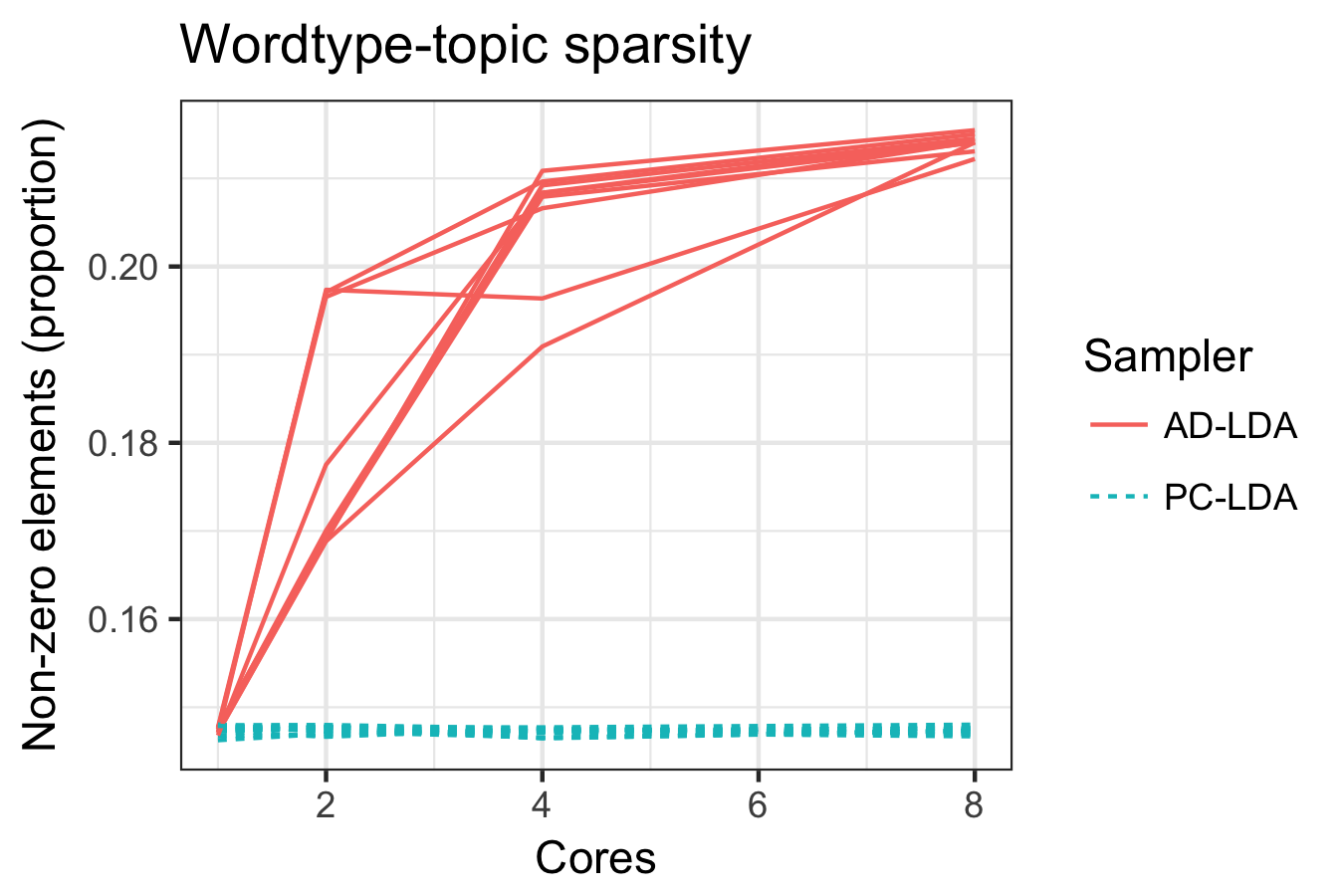}
\includegraphics[scale=0.14]{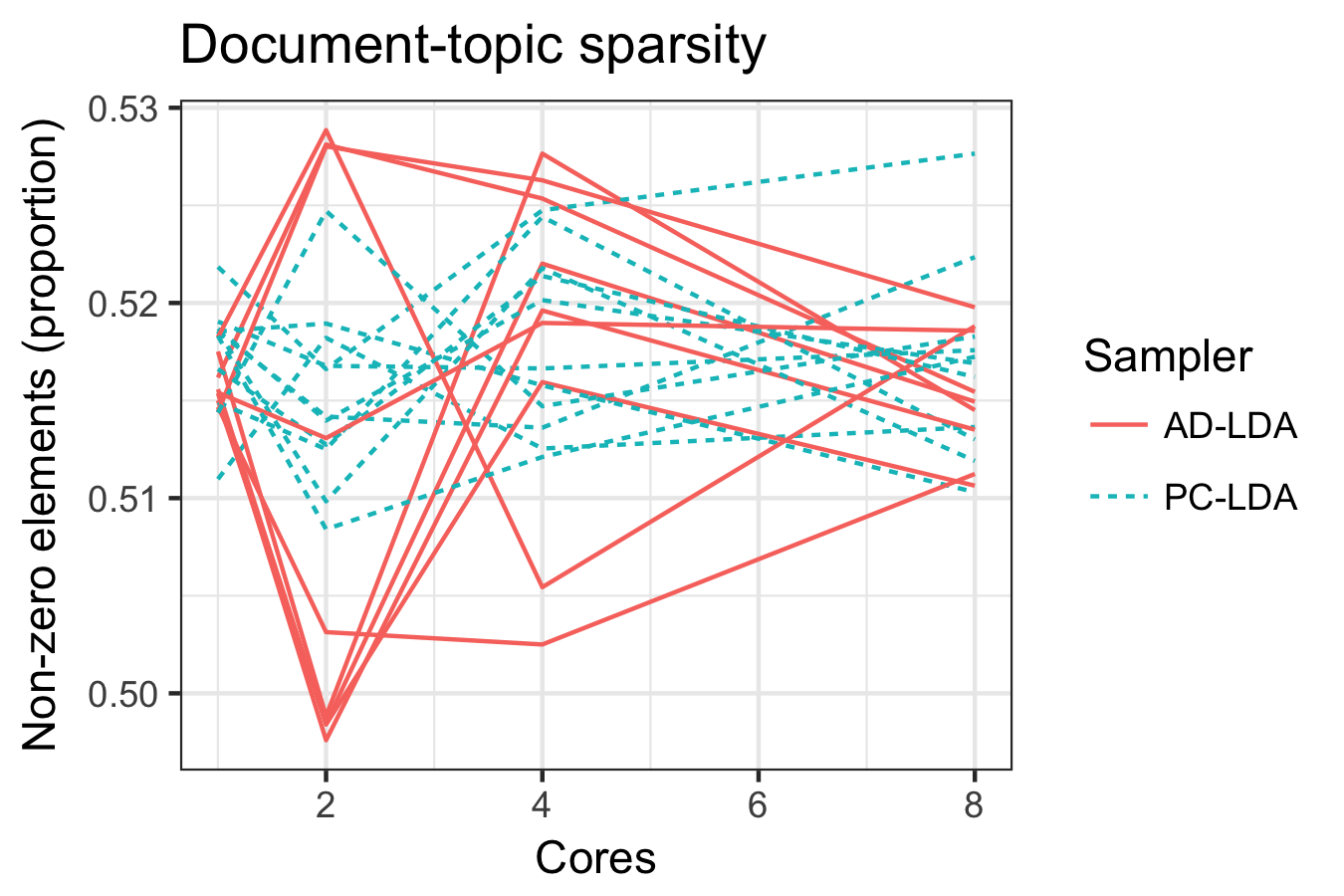} 
\par\end{centering}
\begin{centering}
\includegraphics[scale=0.14]{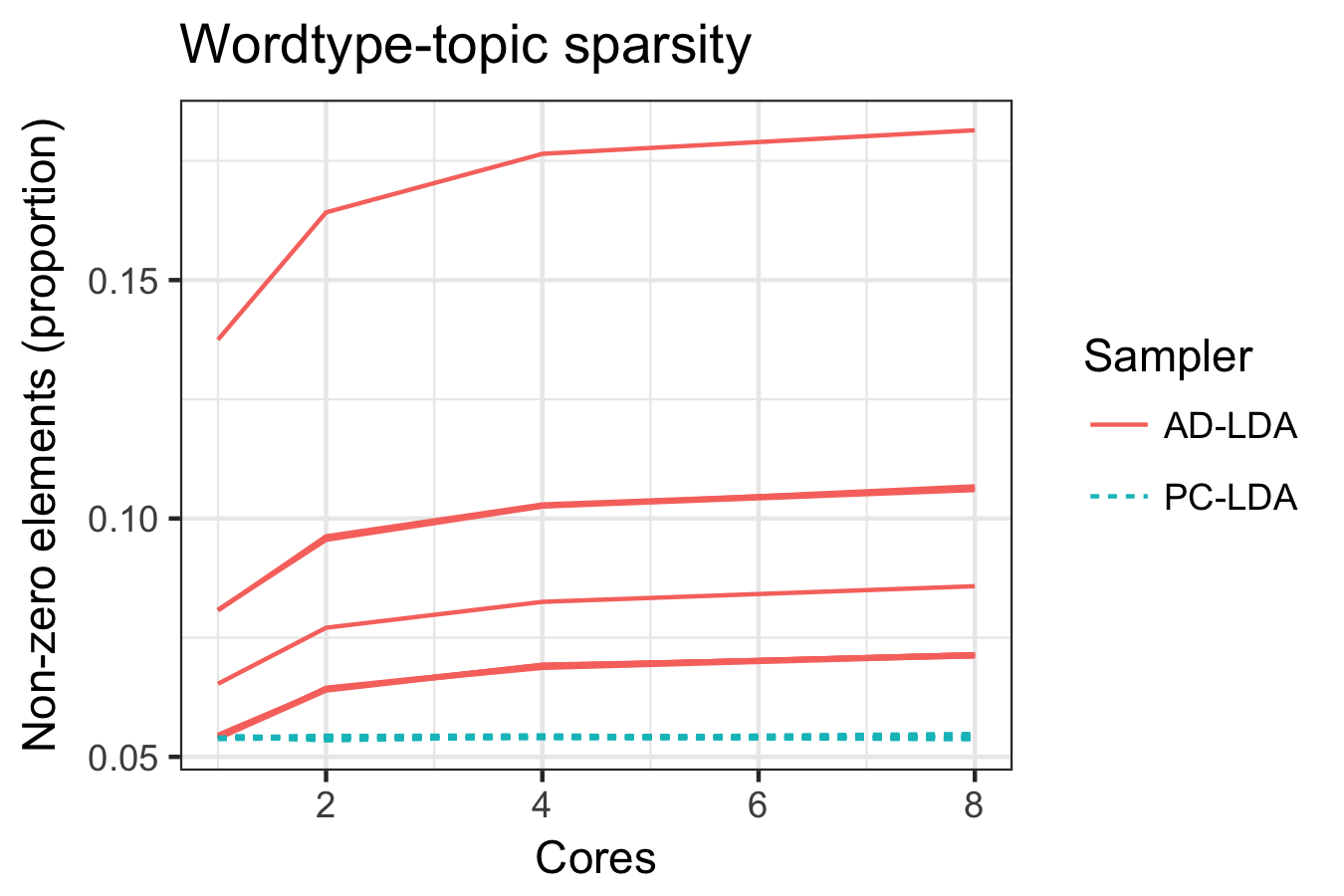}
\includegraphics[scale=0.14]{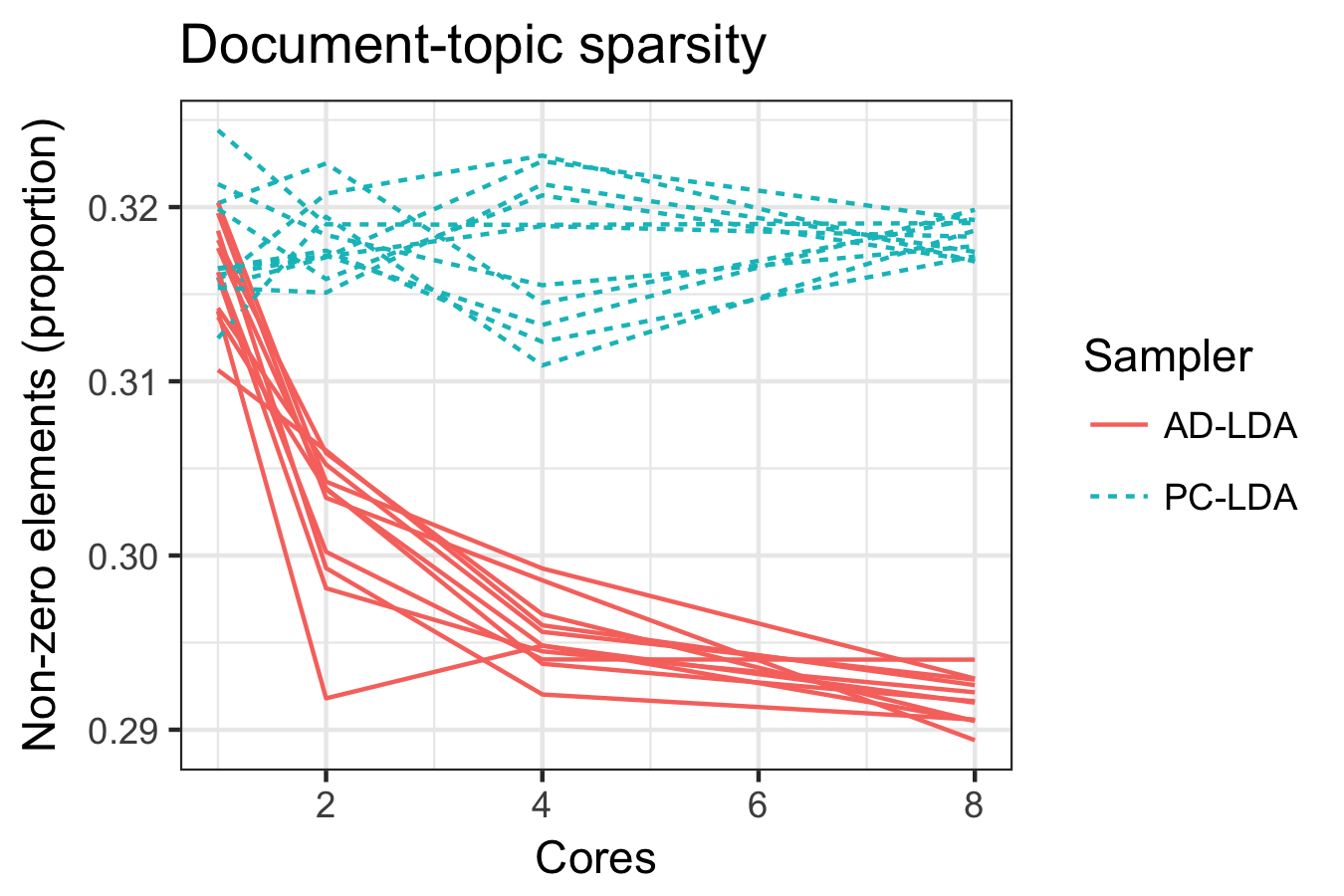}
\par\end{centering}
\vspace{0.3in}
 \caption{The sparsity of $n^{(w)}$ (left) and $n^{(d)}$ (right) as a function
of cores for the NIPS dataset with $K=20$ (upper) and $K=100$ (lower).}
\label{fig:sparsity_nips20} 
\end{figure}

As shown in Figure \ref{fig:loglik_nips20}, there is a clear tendency for AD-LDA to converge to a lower posterior mode as more cores are used to parallelize the sampler. To get some more insights into this behavior of AD-LDA, Figure \ref{fig:sparsity_nips20} displays the sparsity of the $n^{(w)}$ and $n^{(d)}$ matrices (the fraction of elements larger than zero) as a function of the number of cores. This effect is of interest for two reasons. First, this means that AD-LDA does not approximate the posterior with a worse log marginalized posterior, but it approximates the posterior with different properties, and how much the AD-LDA approximation differs with a MCMC approximation depends on the number of cores. Second, we can interpret these results as that using the AD-LDA approximation of the posterior will make the approximate posterior drift towards finding a better local approximation on each core (a more sparse $n^{(d)}$ matrix) and a less good global approximation (a less sparse $n^{(w)}$ matrix). Similar results are found for the Enron corpus (not shown).

The second aspect of the partially collapsed sampler compared with the fully collapsed sampler is that the fully collapsed sampler seems to have a larger problem with getting stuck in local modes when it comes to $n^{(w)}$. As can be seen in Figure \ref{fig:sparsity_nips20}, the sparsity of $n^{(w)}$ for different initial states get stuck at various sparsity levels. In the case of Enron, this happened for one initial seed while in the NIPS 100 situation we can see that the sampler gets stuck at four different sparsity levels. The partially collapsed sampler on the other hand always ends up at the most sparse global solution. This result seems to indicate that the partially collapsed sampler is more robust to initial states and the number of cores used. It would be interesting to follow up these empirical observations by a careful theoretical analysis, but that is beyond the scope of this paper.

\subsection{Parallelism and execution time comparison\label{subsec:Speedup-and-parallelism}}

We compare our proposed samplers with two state-of-the-art samplers: sparse LDA (parallelized using AD-LDA) by \citet{Yao2009} using the original implementation in Mallet and light-LDA by \citet{yuan2015lightlda}, which we implemented in Mallet. Having implemented all samplers in the same Mallet framework makes for a fair comparison between the samplers. There are still differences in that the work of \citet{yuan2015lightlda} has focused on the distributed setting rather than a multicore shared-memory setting. We have chosen to not compare with Alias-LDA since it is similar to light-LDA in that it uses a conditional Metropolis-Hastings approach, but light-LDA has been shown to be faster \citep{yuan2015lightlda}. The samplers are compared using 10, 100, and 1000 topics for the full (100\%) PubMed corpus and a subset (10\%) of the corpus. As explained at the beginning of Section \ref{sec:Experiments}, we compare the samplers in how quickly they reach a region of high posterior density. We refer to this as the speed to reach a mode region.

\begin{figure}[h]
\vspace{0.3in}
 
\begin{centering}
\includegraphics[scale=0.14]{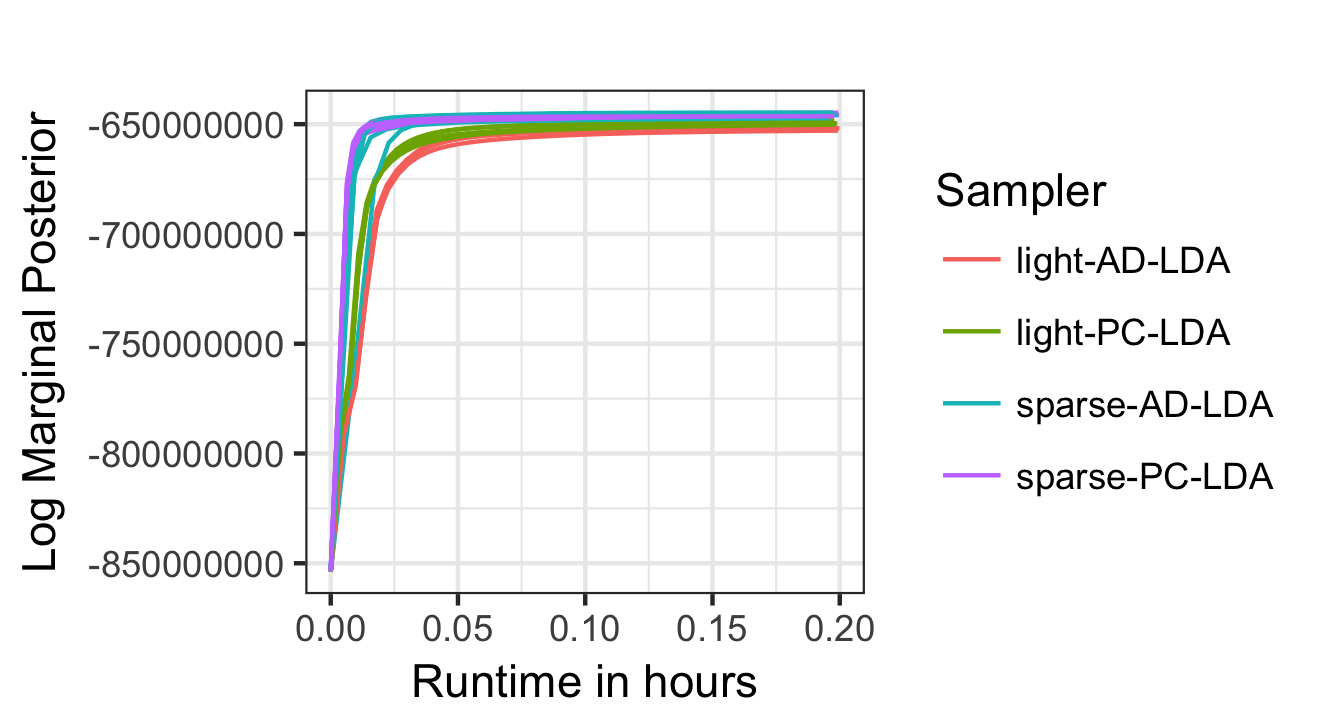}
\includegraphics[scale=0.14]{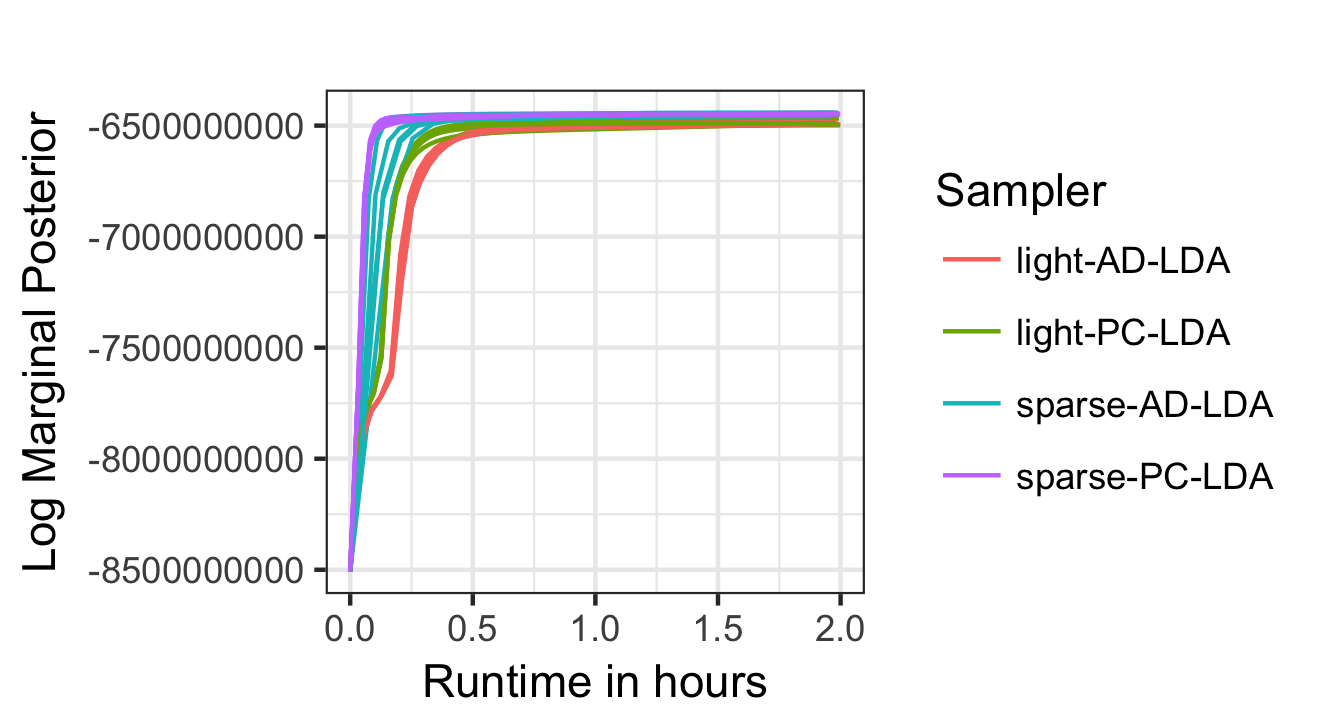}
\par\end{centering}
\begin{centering}
\includegraphics[scale=0.14]{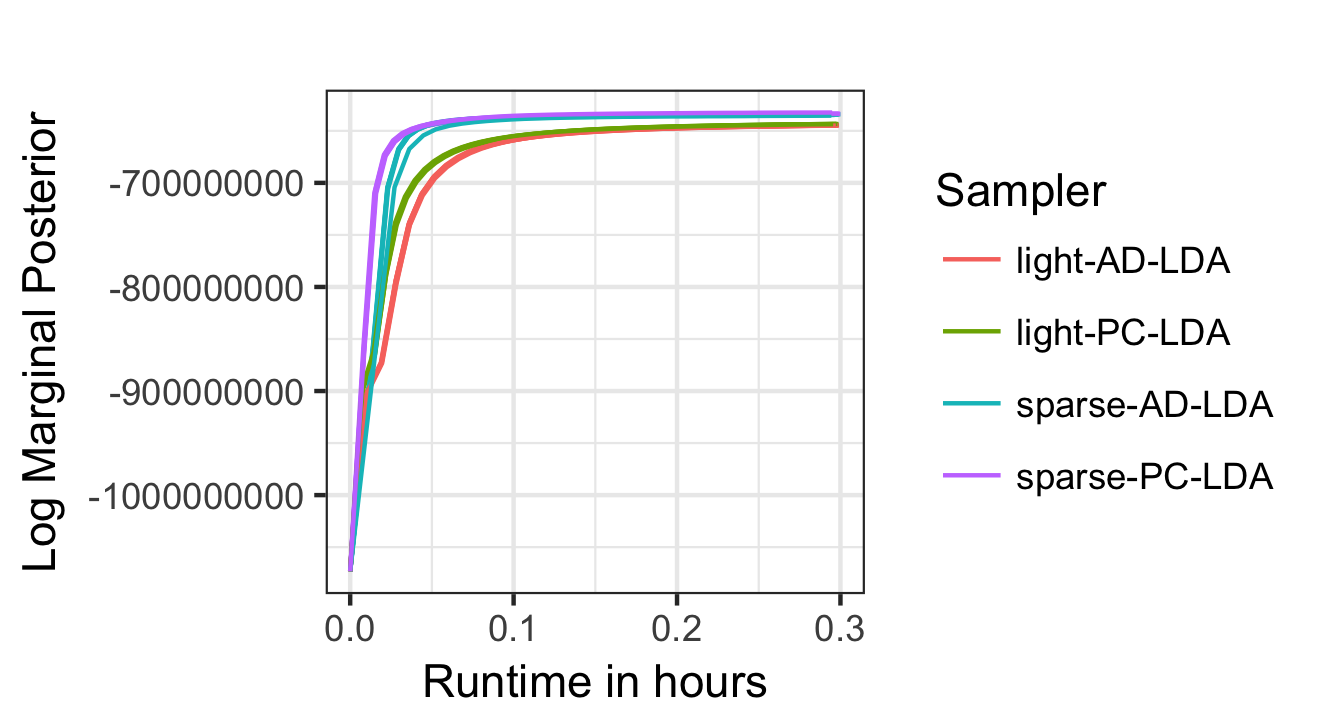}
\includegraphics[scale=0.14]{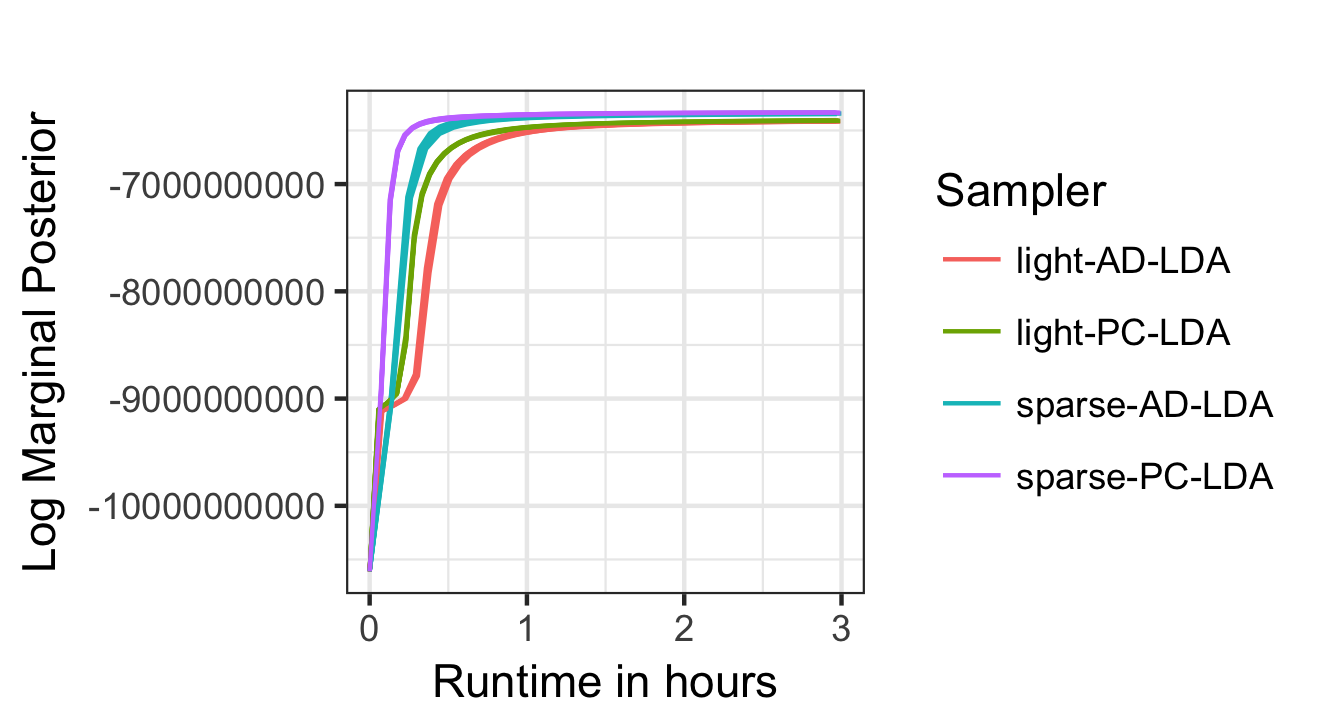}
\par\end{centering}
\begin{centering}
\includegraphics[scale=0.14]{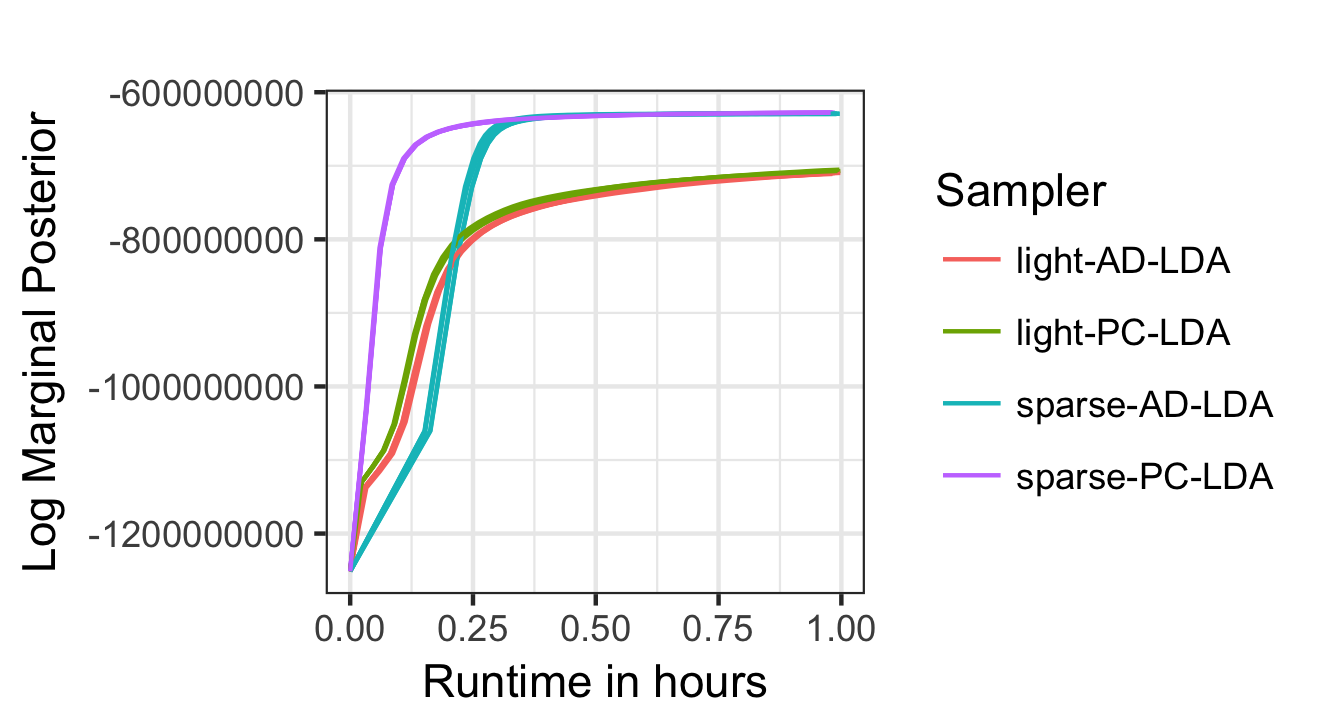}
\includegraphics[scale=0.14]{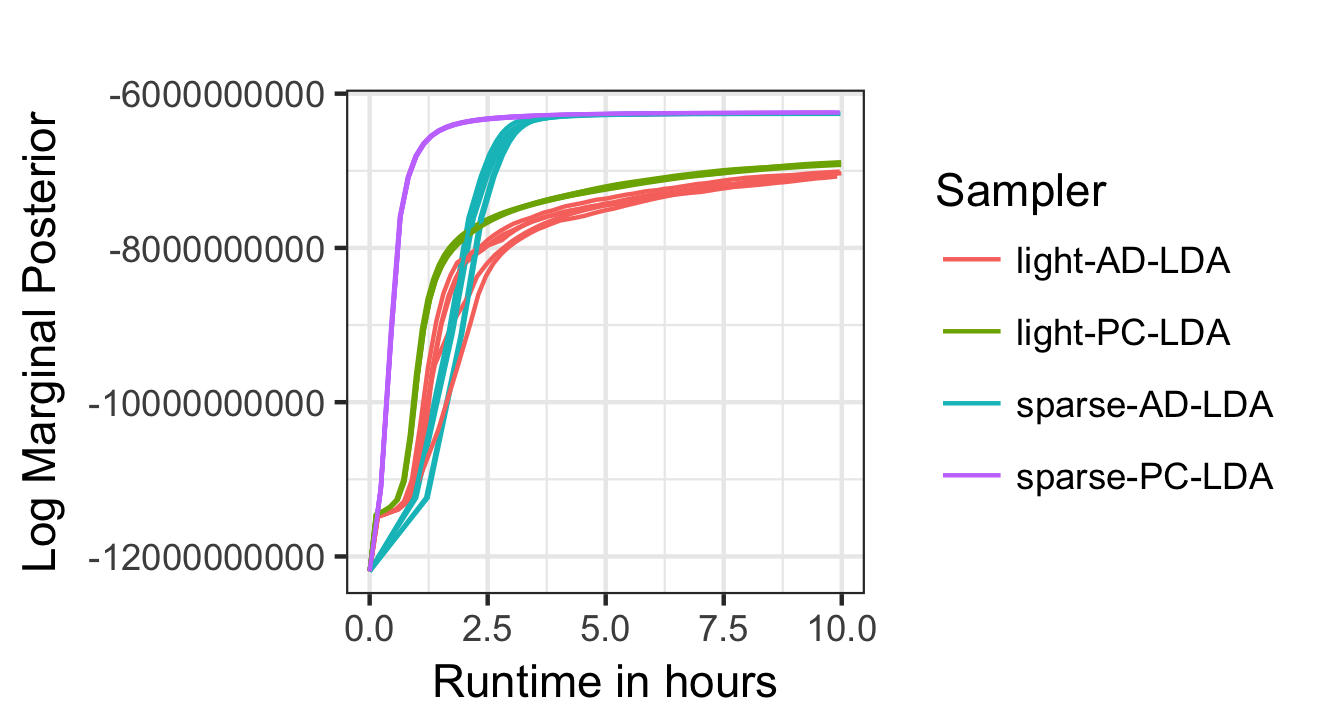}
\par\end{centering}
\vspace{0.3in}
 \caption{Log marginal posterior by runtime for PubMed 10\% (left) and PubMed
100\% (right) for 10, 100, and 1000 topics using 16 cores and 5 different
random seeds.}
\label{fig:pubmed_algo} 
\end{figure}

Figure \ref{fig:pubmed_algo} shows that PC-LDA is in general faster to reach the mode region than most other approaches, especially when the number of topics is large. The pattern is very similar for all corpora sizes. The different light-LDA approaches are increasing fast in log marginal posterior in the initial iterations when sparse-LDA still is working with a more dense matrix, making light-LDA quicker in the beginning. \citet{yuan2015lightlda} show that light-LDA outperforms both sparse LDA and Alias-LDA, a result that differs from our results. We believe that this may be due to implementation details (after personal correspondence with Jinhui Yuan). \citet{yuan2015lightlda} work with a distributed, multi-machine approach while we have done the implementations in a shared memory, multi-core setting. The shared-memory situation is relevant for many practitioners working with larger corpora.

Light-LDA and similar approaches have been shown to work very well for a large number of topics. A very large number of topics may be needed for web-size applications like the proprietary Bing corpus, whereas a much more moderate number of topics is likely to be more useful in less extreme situations. For example, \citet{hoffman2013stochastic} find that a surprisingly small number of topics are optimal in several relatively large corpora of interest for practitioners. As a comparison, we evaluate the speed to the mode region of our samplers using the same settings as in \citet{hoffman2013stochastic} on their Wikipedia corpus \footnote{We could not find the exact same Wikipedia corpus and used a smaller Wikipedia corpus. We still used 100 topics.} (using 7,700 word types) and New York Times corpus (using 8,000 word types). \citet{hoffman2013stochastic} conclude that 100 topics are the optimal number of topics for both corpora.

\begin{figure}[h]
\vspace{0.3in}
 
\begin{centering}
\includegraphics[scale=0.14]{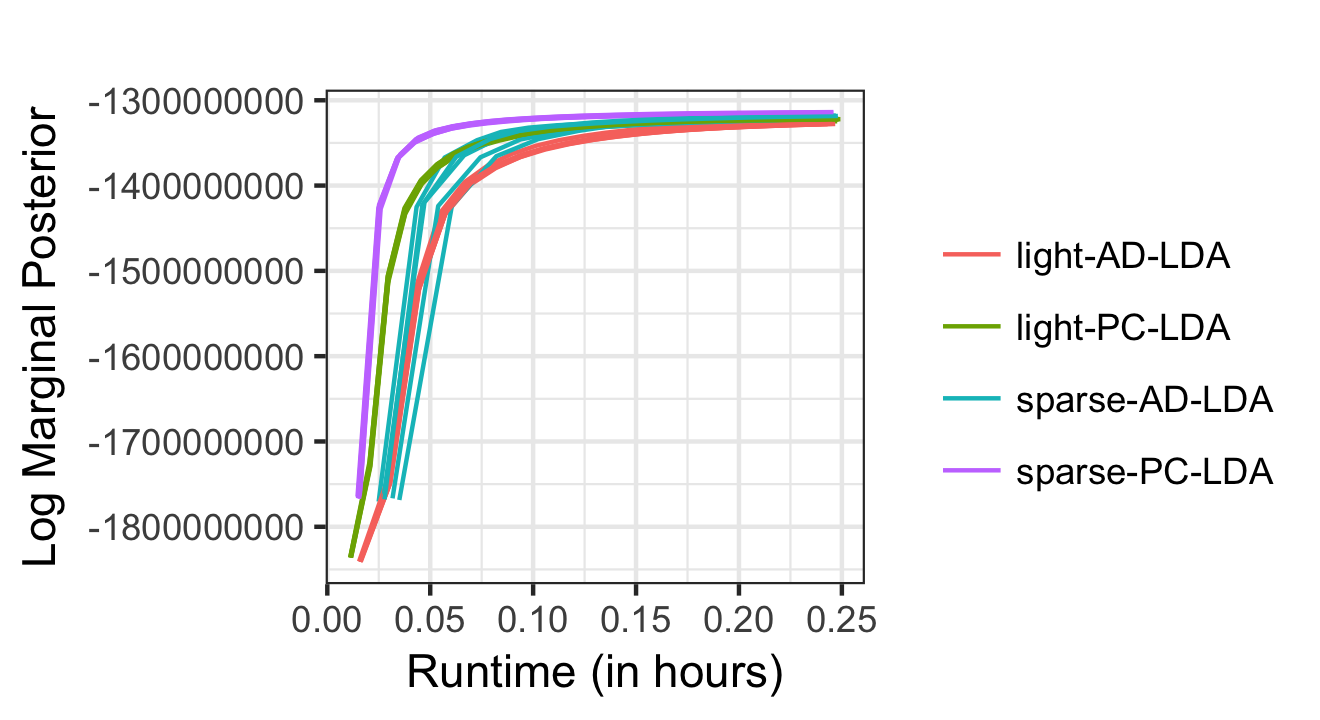}
\includegraphics[scale=0.14]{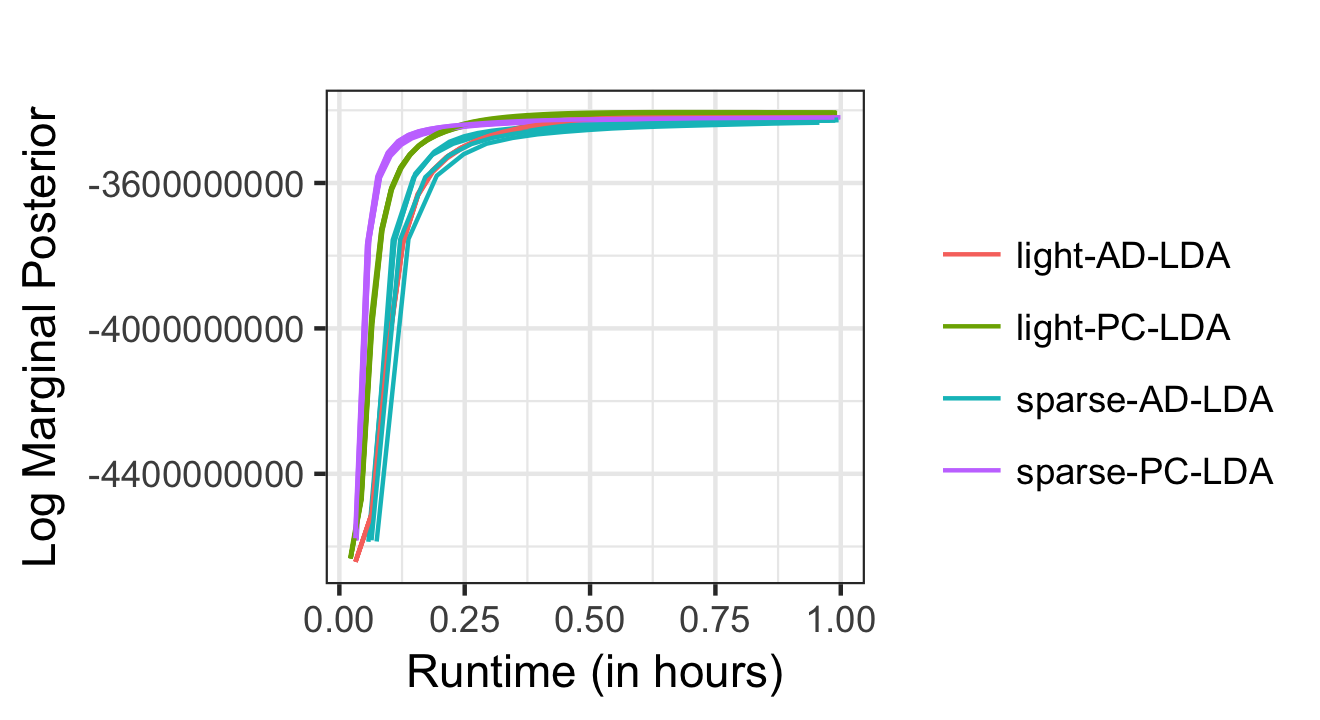}
\par\end{centering}
\vspace{0.3in}
 \caption{Log marginal posterior by runtime for Wikipedia corpus (left) and the New York Times corpus (right) for 100 topics using 16 cores.}
\label{fig:nytwiki_speed} 
\end{figure}

As can be seen in Figure \ref{fig:nytwiki_speed}, most algorithms work well and can fit these models with speeds comparable to that of \citet{hoffman2013stochastic}, using a 16 core machine. Since \citet{hoffman2013stochastic} uses stochastic Variational Bayes to approximate the posterior, the speed of our provably correct MCMC samplers is quite impressive. We can also conclude that although the algorithms are quite similar in speed, the PC-LDA sampler is the winner when it comes to quickly finding the high-density region of the posterior distribution.

To study how the samplers scale as we increase the number of topics, we run PC-LDA for 1000 iterations on the larger PubMed corpus using 100 and 1000 topics and compare the speed until reaching the mode region on 16, 32, and, 64 cores. 

\begin{figure}[h]
\vspace{0.3in}
 
\begin{centering}
\includegraphics[scale=0.14]{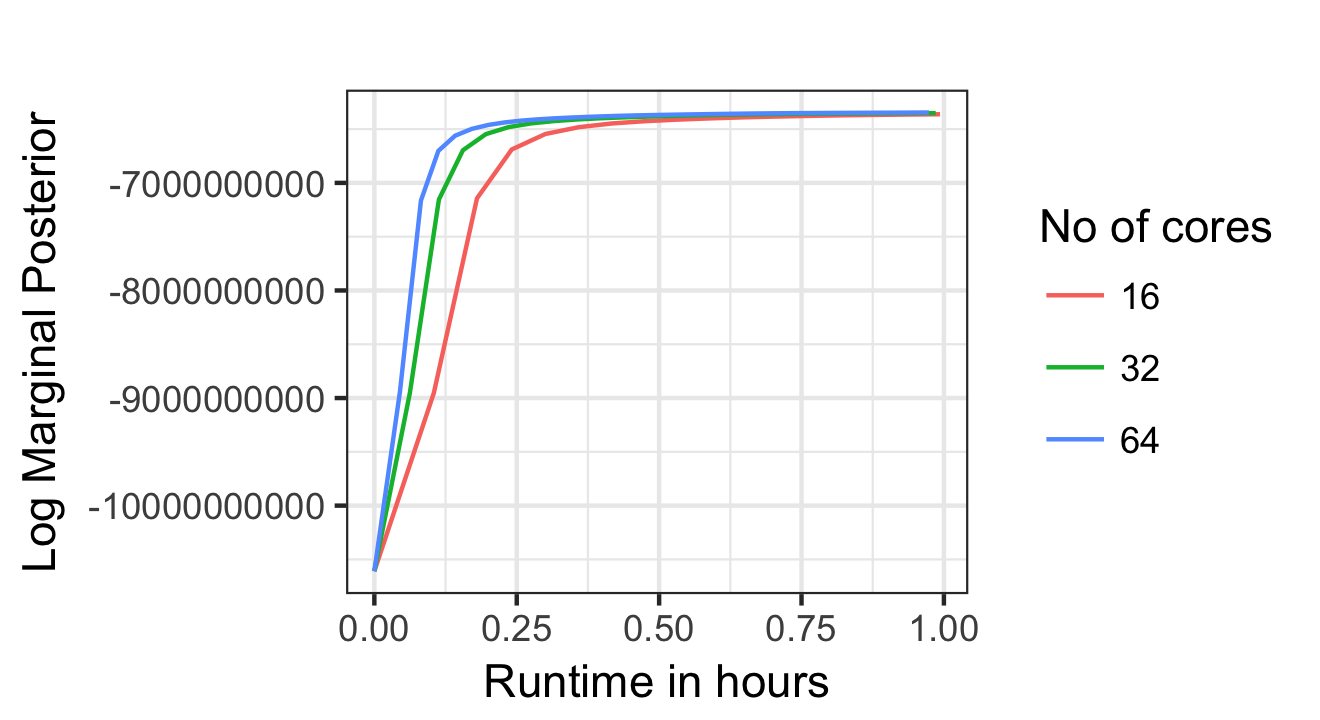}
\includegraphics[scale=0.14]{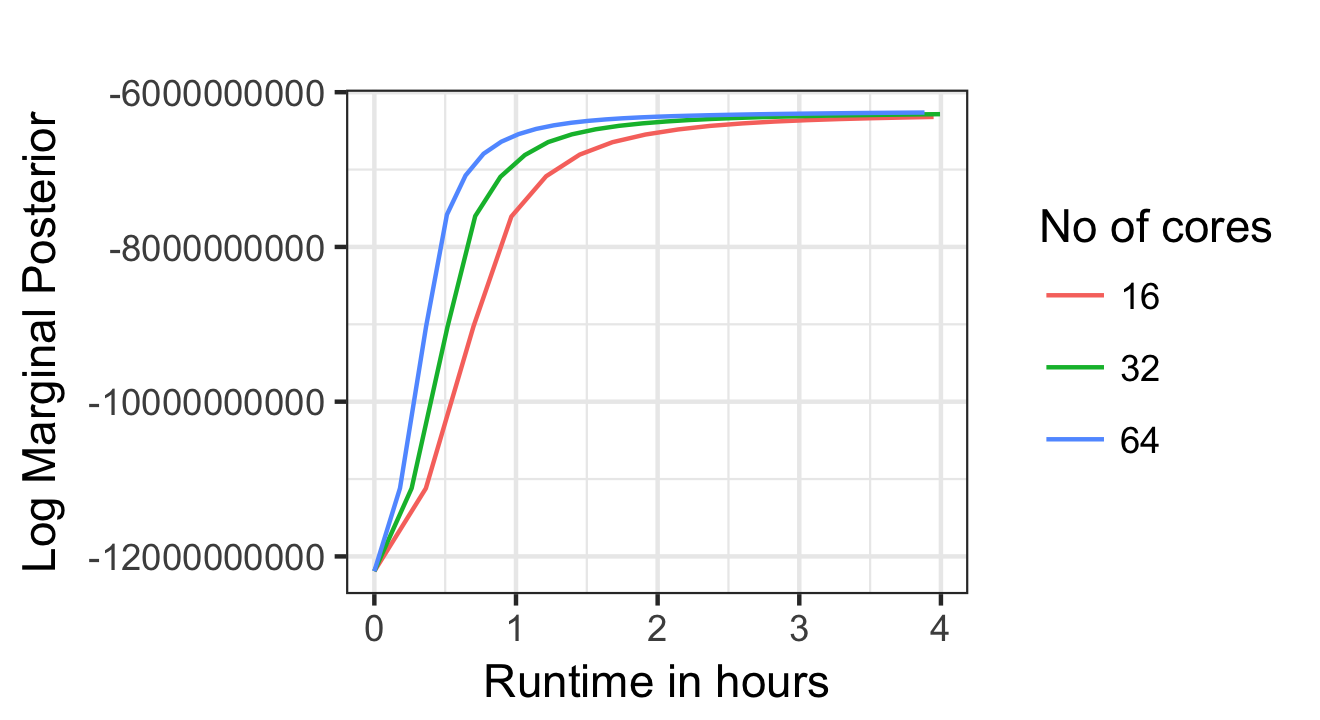} 
\par\end{centering}
\vspace{0.3in}
 \caption{Log marginal posterior by runtime for the PubMed corpus for 100 topics
(left) and 1000 topics (right) using PC-LDA.\label{fig:pubmed}}
\end{figure}

From the data in Figure \ref{fig:pubmed}, we calculate the time it takes for the PC-LDA sampler to reach the mode region (where this
region is defined as 1 \% of the top log marginalized posterior for the sampler for the respective number of topics). The results are presented in Table \ref{tab:execution-times}. By increasing the number of cores from 16 to 64, we can reduce the sampling time to reach the high-density region by 50\%. 

\begin{table}[h]
\caption{Runtime to reach the high density region for 100 and 1000 topics on
16, 32, and 64 cores.}
\label{tab:execution-times} 
\centering{}%
\begin{tabular}{llr}
\textbf{No. Cores}  & \textbf{K}  & \textbf{Runtime (min)} \\
\hline &  & \\
16 & 100 & 49.2\\
32 & 100  & 35.5\\
64 & 100 & 27.5\\
16 & 1000 & 290\\
32 & 1000 & 201\\
64 & 1000 & 141\\
\end{tabular}
\end{table}

Table \ref{tab:execution-times} also shows that PC-LDA makes it possible to reach the interesting part of the posterior in an LDA model with 1000 topics for the large PubMed corpus in a little more than 2 hours on a 64 core machine. 

\section{Discussion and conclusions}

We propose PC-LDA, a sparse partially collapsed Gibbs sampler for LDA. Contrary to state-of-the-art parallel samplers, such as AD-LDA, our sampler is guaranteed to converge to the true posterior. This guarantee is an important property as our experiments indicate that AD-LDA does not converge to the true posterior. This error seems to increase with the number of cores. Although the differences may be small in practice for the basic LDA model, they may very well be amplified in more complicated models or online approaches.

Our PC-LDA sampler is shown under reasonable assumptions to have the same complexity, $O\left(\sum_{i=1}^{N}K_{d(i)}\right)$, per iteration as other efficient collapsed sparse samplers such as Alias-LDA, despite the additional sampling of $\Phi$. The light-PC-LDA sampler is proved to have the same complexity as light-LDA, $O\left(N\right)$ per iteration. The reduced computational complexities of light-PC-LDA and light-LDA do not compensate for the decrease in sampling efficiency and the PC-LDA sampler with direct sampling from $p(\mathbf{z}\vert\Phi)$ is, in general, the sampler which most quickly reaches the region around the dominant mode in our experiments. 

An effective sampler for topic models needs to balance three entities in an optimal way: sampling complexity, sampling efficiency, and constant factors; the time complexity of the sampler is important but is not the whole story. All Metropolis-Hastings samplers presented here are of complexity $O(N)$, but this reduction in sampling complexity comes at the cost of reduced sampling efficiency. Light-LDA trades off the mixing efficiency of the chain to reduce the sampling complexity and PC-LDA trades off efficiency to enable a parallel sampler that converges to the true posterior. Lastly, even for large corpora, the constant factors are important. PC-LDA needs to sample $\Phi$, and Light-LDA needs to draw multiple random variates per sampled topic indicator. We have for example seen that sparse samplers are having more difficulties in distributing the sampling workload between cores than the light samplers. 

\citet{yuan2015lightlda} show that light-LDA outperforms both sparse LDA and Alias-LDA, a result that differs from our results. We believe that this may be due to implementation details (after personal correspondence). \citet{yuan2015lightlda} work with a distributed, multi-machine approach while we have done the implementations in a shared memory, multi-core setting. The shared-memory situation is relevant for many practitioners working with larger corpora.

We believe that the reason for the success of PC-LDA is three-fold:
First, PC-LDA (like Alias-LDA) limits the sampling complexity to the number of topics in each document, which tends to be small in practice. Assuming that the number of tokens in the documents is finite, this complexity can be regarded as constant since the complexity is limited by the document size even when $K\rightarrow\infty$. 
Second, PC-LDA (unlike light-PC-LDA, light-LDA, and Alias-LDA) is a Gibbs sampler that only needs to draw one random variable per token and iteration. Drawing random variates are relatively costly, so even if the computational complexity is reduced using light-PC-LDA or light-LDA, the constant cost of sampling a single token is larger. 
Third, contrary to common belief, we demonstrate on several commonly used corpora that a partially collapsed sampler has nearly the same MCMC efficiency as the gold standard sequential collapsed sampler, but enjoys the advantage of straightforward parallelization. Our results show speedups per iteration to at least 64 cores on the larger PubMed corpus, making it a good option for MCMC sampling in larger corpora.

An important advantage of PC-LDA is that it also allows for more interesting non-conjugate models for $\Phi$, such as regularized topic models \citep{newman2011improving} or using a distributed stochastic gradient MCMC approach as has been proposed in \citet{ahn2014distributed}. As an example of such an extension, Appendix \ref{subsec:Variable-selection-description} gives the details for a PC-LDA algorithm that allows for variable selection in the topics, where elements in $\Phi$ may be set to zero.

In summary, we propose and evaluate new sparse partially collapsed Gibbs samplers for LDA with several algorithmic improvements. Our preferred algorithm, PC-LDA, is fast, efficient, and exact. Compared to the popular collapsed AD-LDA sampler, PC-LDA is applicable to a larger class of extended LDA models as well as in other language models using the Dirichlet-multinomial conjugacy.

\section*{Acknowledgements}

This research is in part financially supported by the Swedish Foundation for Strategic Research (SSF) (project ASSEMBLE, RIT15-0012 and project Smart Systems: RIT 15-0097).

\bibliographystyle{plainnat}
\bibliography{references}

\appendix

\section{Proofs \label{subsec:Proofs}}

\paragraph*{Complexity of the sparse partially collapsed Gibbs sampler\label{Prop1-1}}

\textit{Assuming a vocabulary size following Heaps' law $V=\xi N^{\varphi}$
with $\xi>0,0<\varphi<1$ and the number of topics following the mean
of a Dirichlet process mixture $K=\gamma\log\left(1+\frac{N}{\gamma}\right)$
with $\gamma>0$, the complexity of the PC-LDA sampler is} 
\[
O\left(\sum_{i=1}^{N}K_{d(i)}\right).
\]

\begin{proof} Under the assumptions in the proposition, the complexity
of PC-LDA for large $N$ is
\[
\sum_{i}^{N}K_{d(i)}+\gamma\xi\log\left(1+\frac{N}{\gamma}\right)N^{\varphi}.
\]
We, therefore, want to prove that there exists a $c>0$ and a $N_{0}\leq N$
such that 
\begin{eqnarray*}
\sum_{i}^{N}K_{d(i)}+\gamma\xi\log\left(1+\frac{N}{\gamma}\right)N^{\varphi} & \leq & c\sum_{i}^{N}K_{d(i)},
\end{eqnarray*}
or, equivalently, 
\[
1+\gamma\xi\log\left(1+\frac{N}{\gamma}\right)N^{\varphi}/\sum_{i}^{N}K_{d(i)}\leq c,
\]
for all $N\geq N_{0}$. Since $N\leq\sum_{i}^{N}K_{d(i)}$ we have
\begin{eqnarray*}
1+\gamma\xi\log\left(1+\frac{N}{\gamma}\right)N^{\varphi}/\sum_{i}^{N}K_{d(i)} & \leq & 1+\gamma\xi\log\left(1+\frac{N}{\gamma}\right)N^{\varphi}/N\:.
\end{eqnarray*}
It is, therefore, enough to show that for $N\geq N_{0}=1$, there
exist a $c$ such that 
\[
1+\gamma\xi\log\left(1+\frac{N}{\gamma}\right)N^{\varphi}/N\leq c\:.
\]
Let 
\[
f(N)=\log\left(1+\frac{N}{\gamma}\right)/N^{1-\varphi}\,,
\]
and note that 
\[
f(1)=\log(1+\gamma^{-1})>0\mbox{ and }\lim_{N\rightarrow\infty}f(N)=0
\]
by the standard limit 
\[
\lim_{x\rightarrow\infty}\left(\frac{\log(x)}{x^{b}}\right)=0\mbox{ for }b>0.
\]
There exist an $R$ such that for all $N>R,f(N)<f(1)$. Using the
extreme value theorem we know that at the interval $[1,R]$ there
exist a $M=\sup f(N)$. Hence, for $N\geq N_{0}=1$, there exists
a $c$ such that $c=1+\gamma\xi M$, which completes the proof.\end{proof}

\paragraph*{Proposition 2. Complexity of the light partially collapsed conditional
Metropolis-Hastings sampler\label{Prop2-1}}

\textit{Assuming a vocabulary size following Heaps' law $V=\xi N^{\varphi}$
with $\xi>0,0<\varphi<1$ and the number of topics following the mean
of a Dirichlet process mixture $K=\gamma\log\left(1+\frac{N}{\gamma}\right)$
with $\gamma>0$, the complexity for the light-PC-LDA sampler is}
\[
O\left(N\right).
\]

\begin{proof} Under the assumptions in the proposition, the complexity
of light-PC-LDA for large $N$ is
\[
N+\gamma\xi\log\left(1+\frac{N}{\gamma}\right)N^{\varphi}\:,
\]
and we, therefore, need to prove that the exists a $c>0$ and $N_{0}\leq N$
such that 
\begin{eqnarray*}
N+\gamma\xi\log\left(1+\frac{N}{\gamma}\right)N^{\varphi} & \leq & cN
\end{eqnarray*}
or equivalently
\[
1+\gamma\xi\log\left(1+\frac{N}{\gamma}\right)N^{\varphi}/N\leq c
\]
for all $N\geq N_{0}$. The rest of the proof follows the proof of
Proposition 1 exactly.
\end{proof}

\section{Variable selection in $\Phi$ \label{subsec:Variable-selection-description}}

Partially collapsed sampling of topic models has the additional advantage that more complex models can be used to model $\Phi$. As an example, we derive a Gibbs sampler for a topic model with a spike-and-slab type prior for $\Phi$ that assigns point masses at zero to a subset of the parameters in $\Phi$. Variable selection for LDA has previously been proposed by \citet{chien2014bayesian} using variational Bayes inference where the sparsity reduced both perplexity as well as memory and computation costs; we will here derive a similar approach using Gibbs sampling.

The rows of $\Phi$ are assumed to be independent a priori, exactly like in the original LDA model, so let us focus on a given row $\phi_{k}$ of $\Phi$. Let $n_{k}^{(w)}$ be the $k$th row of $n^{(w)}$, and let $I_{k}=(I_{k1},...,I_{kV})$ be a vector of binary variable selection indicators such that $I_{k,v}=1$ if $\phi_{k,v}>0$, and $I_{k,v}=0$ if $\phi_{k,v}=0$ where $v$ is the word type (column) of the $\Phi$ matrix. Define $I_{k}^{c}$ to be the complement of $I_{k}$ (i.e., the vector indicating the zeros in $\phi_{k}$). Let $\phi_{k,I_{k}}$ be a vector with elements of $\phi_{k}>0$. Finally, let $n_{k}$ be the total number of tokens associated with the $k$th topic. The indicators $I_{kv}$ are assumed to be iid $\mbox{Bern}(\pi_{k})$ a priori. The prior for $\phi_{k}$ is a conditional Dirichlet distribution where 

\[
p(\phi_{k,I_{k}}\vert I_{k})\sim\mbox{Dir}(\beta_{I_{k}})
\]

and $\phi_{k,I_{k}^{c}}=0$ with probability one.

The posterior sampling step for $\phi_{k}$ is replaced by two sampling steps. First we sample the indicator variables for each word type $p(I_{k,v}\mid I_{k,-v},\pi_{k},v,\mathbf{z})$ and then, conditional on the indicators $I_{k}$ we sample $p(\phi_{k,I_{k}}\vert I_{k}\mathbf{z})$ from the conditional Dirichlet distribution in \citet{ng2011dirichlet}.

\subsection*{Sampling $\phi_{k}$.}

Sampling from $p(\phi_{k}|I_{k},\mathbf{z})$ is straightforward by setting $\phi_{k,I_{k}^{c}}=0$ and drawing the non-zero elements in $\phi_{k}$ from 

\[
p(\phi_{k,I_{k}}|I_{k},\mathbf{z})\sim\mbox{Dir}(\beta_{I_{k}}+n_{I_{k}}^{(w)})\:.
\]

\subsection*{Sampling $I_{k,v}$. }

We first note that if $n_{k,v}^{(w)}>0$ we know that $\phi_{k,v}>0$ and hence we can set $I_{k,v}=1$ with probability 1. The conditional posterior distribution of $I_{k,v}$ is a two-point distribution and hence we need to compute the conditional distribution $p(I_{k,v}=1|I_{k,-v},\pi_{k},v,\mathbf{z})$ and $p(I_{k,v}=0|I_{k,-v},\pi_{k},v,\mathbf{z})$ by integrating out $\phi_{k}$. Note that $p(I_{k}|\mathbf{z})=p(I_{k}|n_{k}^{(w)})$ since $n^{(w)}$ is a sufficient statistic for $I_{k}$. By Bayes theorem we get 

\[
p(I_{k}|n_{k}^{(w)})\propto p(n_{k}^{(w)}|I_{k})p(I_{k})\:,
\]

where $p(I_{k})$ is the Bernoulli prior and 

\begin{eqnarray*}
p(n_{k}^{(w)}|I_{k}) & \propto & \int\phi_{k,I_{k}}^{n_{k,I_{k}}^{(w)}+\beta_{I_{k}}-1}d\phi_{k,I_{k}}\\
 & = & \mathbf{B}(n_{k,I_{k}}^{(w)}+\beta_{I_{k}})\:,
\end{eqnarray*}
using the Dirichlet kernel and with $\mathbf{B}$ being the multivariate Beta function. With some algebra we then have that the conditional two-point distribution where we set $I_{k,v}=1$ if $n_{k,v}^{(w)}>0$ and otherwise we draw $I_{k,v}$ using the following two-point distribution:

\begin{eqnarray*}
p\left(I_{k,v}=1\vert I_{k,(-v)},\pi_{k}\right) & \propto & \frac{\Gamma\bigl(\sum_{j\in I_{k},j\neq v}\beta_{k,j}+\beta_{k,v}\bigr)}{\Gamma\bigl(\sum_{j\in I_{k},j\neq v}n_{k,j}^{(w)}+\beta_{k,j}+\beta_{k,v}\bigr)}\pi_{k}\:,\\
p\left(I_{k,v}=0\vert I_{k,-v},\pi_{k}\right) & \propto & \frac{\Gamma\bigl(\sum_{j\in I_{k},j\neq v}\beta_{k,j}\bigr)}{\Gamma\bigl(\sum_{j\in I_{k},j\neq v}n_{k,j}^{(w)}+\beta_{k,j}\bigr)}(1-\pi_{k})\:.
\end{eqnarray*}
where $\Gamma(\cdot)$ is the Gamma function. 

\begin{figure}
\begin{centering}
\includegraphics[scale=0.14]{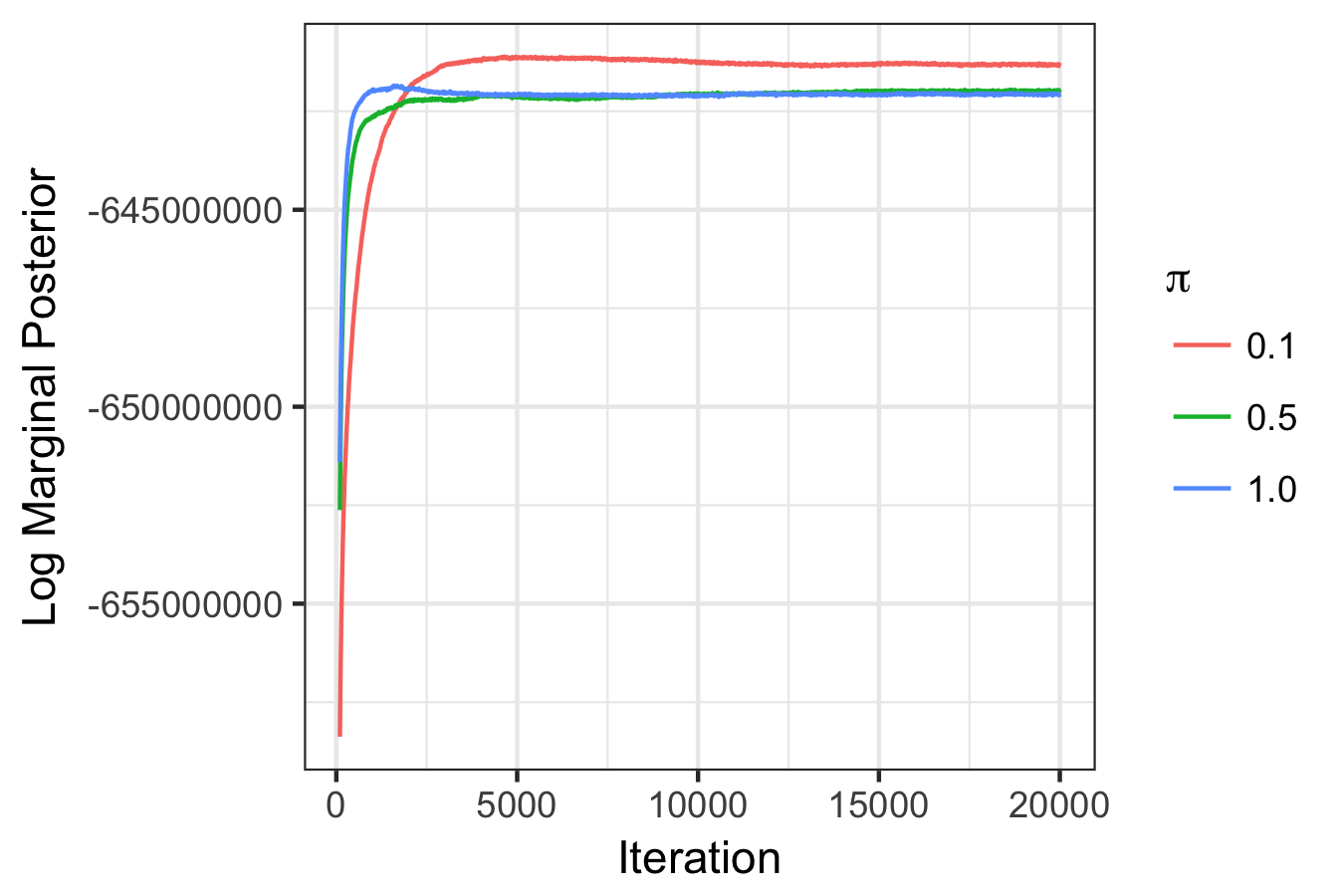}
\includegraphics[scale=0.14]{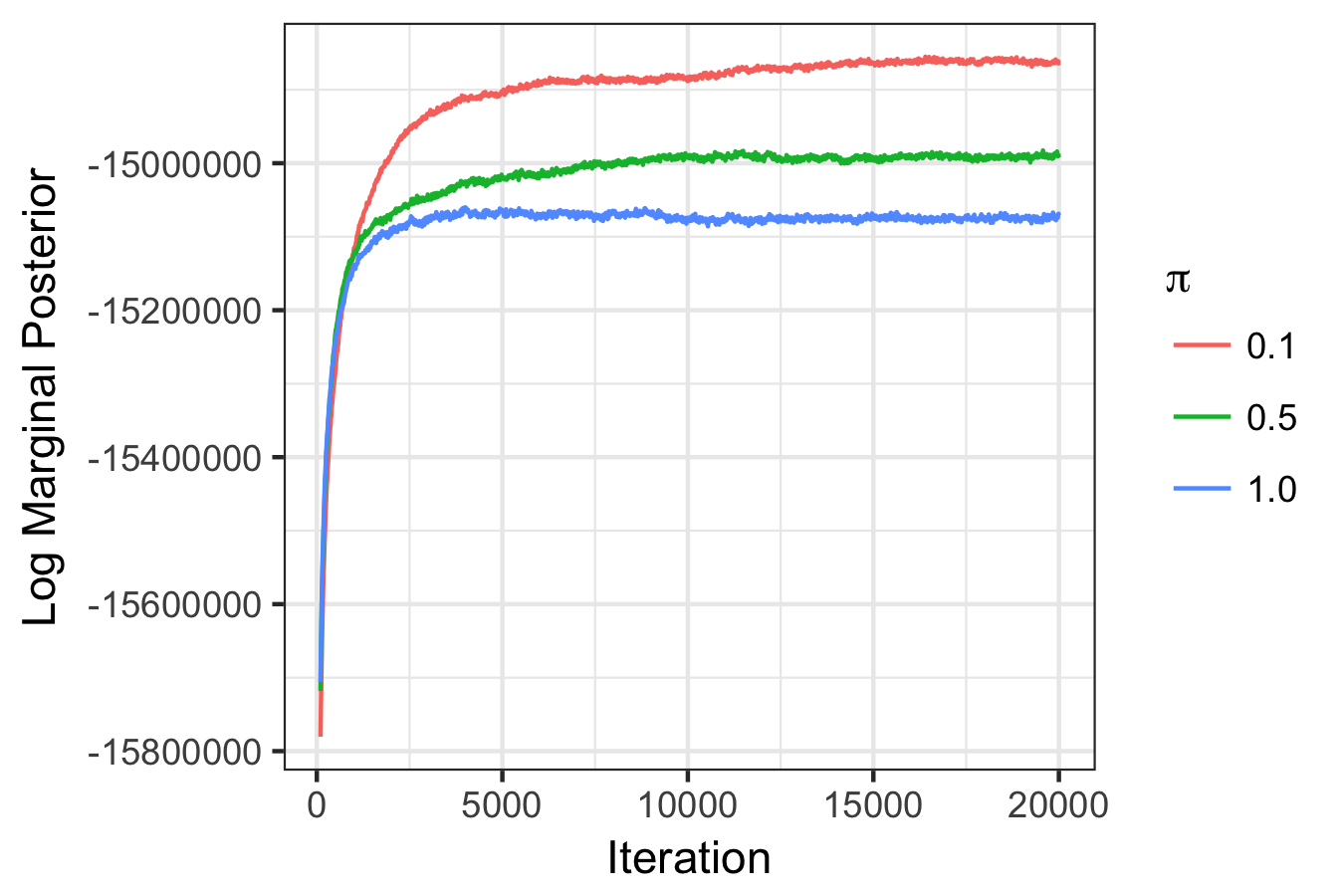}
\label{fig:VariableSelectionPi05-1}
\par\end{centering}
\caption{Log marginalized posterior for different values of $\pi$ for PubMed
10\% (left) and NIPS (right).}
\end{figure}

Earlier research has already concluded that introducing variable selection for $\Phi$ in LDA can reduce the perplexity and increase parsimony of the topic model \citep{chien2014bayesian}. Here we illustrate the effect of variable selection for the PubMed (10\%) and NIPS corpora (both with a rare word limit of 10). We set the sparsity prior to $\pi=1.0,0.5,0.1$, $K=100$, and run all models for 20 000 iterations. We examine the proportions of zeroes and log marginalized posterior induced by the variable selection prior $\pi$. The proportion of zeroes in $\Phi$ is estimated using the last 1000 iterations. The results are summarized in Table \ref{tab:VS_results} and Figure \ref{fig:VariableSelectionPi05-1}.

\begin{table}[h]
\caption{Variable selection of $\Phi$}
\label{tab:VS_results} 
\centering{}%
\noindent\begin{minipage}[t]{1\columnwidth}%
\begin{center}
\begin{tabular}{llll}
\textbf{DATA}  & \textbf{K}  & \textbf{$\pi$} & \textbf{Prop. zeros in $\Phi$} \tabularnewline
\hline &  &  & \tabularnewline
PubMed 10\%  & 100  & 0.1 & 0.879\tabularnewline
PubMed 10\% & 100  & 0.5 & 0.492\tabularnewline
PubMed 10\% & 100  & 1.0 & 0.000 \tabularnewline
NIPS  & 100 & 0.1 & 0.877\tabularnewline
NIPS  & 100 & 0.5 & 0.501\tabularnewline
NIPS  & 100  & 1.0 & 0.000 \tabularnewline
\end{tabular}
\par\end{center}%
\end{minipage}
\end{table}

The results are similar to that of \citet{chien2014bayesian} in that a sparse prior will result in a better marginal likelihood of the model. This model can be elaborated further (the most obvious is learning $\pi$), but this is out of the scope for this paper. 

\section{Algorithms \label{subsec:Algorithms}}

\begin{algorithm}
\SetAlgoLined 
\SetKwData{Left}{left} 
\SetKwData{This}{this} 
\SetKwData{Up}{up} 
\SetKwFunction{Union}{Union} 
\SetKwFunction{FindCompress}{FindCompress}
\KwData{tokenized corpus $\mathbf{x}$} 
\KwResult{topic indicators $\mathbf{z}$}

\BlankLine
$\mathbf{z} \leftarrow$ RandomInitZ()\; 
$\mathbf{n}^{(w)} \leftarrow$ SumUpGlobalSufficientStatistics($\mathbf{z}$)\;
\BlankLine
\For{$g \leftarrow 1$ \KwTo num\_iterations}{
   // Sample $\Phi$

   $\Phi \leftarrow$ SampleDirichlet($\mathbf{n}^{(w)}$, $\beta$)\;

   $ \mathbf{A}, \sigma_a \leftarrow $ ConstructAliasTables($\Phi$, $\alpha$)\;
   // Sample $\mathbf{z}$

   $\mathbf{z},\mathbf{n}^{(w)}  \leftarrow$ \texttt{SampleTopicIndicatorsSparse}($\Phi$, $\mathbf{x}$, $\mathbf{z}$, $\sigma_a$, $\mathbf{n}^{(w)}$)\;
} 

\caption{Sparse Partially Collapsed LDA\label{alg:Sparse-Partially-Collapsed}}
\end{algorithm}

\begin{algorithm}
\SetAlgoLined 
\SetKwData{Left}{left}
\SetKwData{This}{this}
\SetKwData{Up}{up} 

\SetKwFunction{Union}{Union}
\SetKwFunction{FindCompress}{FindCompress} 
\SetKwInOut{Input}{input}
\SetKwInOut{Output}{output}
\KwData{$\Phi$, $\mathbf{x}$, $\mathbf{z}$, $\sigma_a$, $\mathbf{n}^{(w)}$} \KwResult{$\mathbf{z}$,$\mathbf{n}^{(w)}$}

\BlankLine
\For{$d \leftarrow 1$ \KwTo $D$ in parallel}{      
   // Compute sorted sparse sufficient statistic for $\theta_d$

   $\mathbf{n}^{(d)} \leftarrow$ SumUpDocumentSparseTopicCounts($\mathbf{z}_d$)\;          
   \For{$i \leftarrow 1$ \KwTo $N_d$}{
      // Remove position $i$ from the sufficient statistics

      $\mathbf{n}_{z_i}^{(d)}, \Delta \mathbf{n}_{z_i,x_i}^{(w)} \leftarrow -1$ \;
      // Compute normalization constant $\sigma_b$ and cumulative sum $\mathbf{s}$ over topics

      $\sigma_b, \mathbf{s}_b  \leftarrow$ ComputeSparseCumSum($\mathbf{n}^{(d)}, \phi_{x_i}$)\;                  
      $u_i \leftarrow$ RandomUniform()\;
      $u_\sigma \leftarrow u_i \cdot (\sigma_a + \sigma_b)$\;
      \eIf{$u_\sigma  < \sigma_a$}{
          // Normalize the random draw to [0,1]

          $u_i \leftarrow u_\sigma / \sigma_a$\;
          // Sample from "prior" part using $u_i$
 
          $z_i  \leftarrow$ \texttt{LookUpAliasTable}($A_{x_i}, u_i$)\;
      }{
         // Normalize the random draw to [0,1]

         $u_i \leftarrow (u_\sigma - \sigma_a) / \sigma_b$ \;
         // Sample from sparse "likelihood" part using $u_i$

        $z_i  \leftarrow$ BinarySearch($\mathbf{s}_b, u_i \cdot \sigma_b $)\;
     } 
      // Add the new topic indicator to the sufficient statistics

      $\mathbf{n}_{z_i}^{(d)}, \Delta \mathbf{n}_{z_i,x_i}^{(w)} \leftarrow +1$ \;
   }     
   $\mathbf{n}^{(w)} \leftarrow$ UpdateGlobalSufficientStatistics($\mathbf{n}^{(w)}$, $\Delta \mathbf{n}^{(w)}$) 
}

\caption{\texttt{SampleTopicIndicatorsSparse()\label{alg:SampleTopicIndicatorsSparse()}}}
\end{algorithm}

\begin{algorithm}
\SetAlgoLined 
\SetKwData{Left}{left} 
\SetKwData{This}{this} 
\SetKwData{Up}{up} 
\SetKwFunction{Union}{Union} 
\SetKwFunction{FindCompress}{FindCompress}
\KwData{tokenized corpus $\mathbf{x}$} 
\KwResult{topic indicators $\mathbf{z}$}

\BlankLine
$\mathbf{z} \leftarrow$ RandomInitZ()\; 
$\mathbf{n}^{(w)} \leftarrow$ SumUpGlobalSufficientStatistics($\mathbf{z}$)\;
\BlankLine
\For{$g \leftarrow 1$ \KwTo num\_iterations}{
   // Sample $\Phi$

   $\Phi \leftarrow$ SampleDirichlet($\mathbf{n}^{(w)}$, $\beta$)\;

  $ \mathbf{A} \leftarrow $ ConstructAliasTables($\Phi$)\;
   // Sample $\mathbf{z}$

   $\mathbf{z},\mathbf{n}^{(w)}  \leftarrow$ \texttt{SampleTopicIndicatorsLight}($\mathbf{A}$, $\mathbf{x}$, $\mathbf{z}$, $\mathbf{n}^{(w)}$)\;
} 

\caption{Light Partially Collapsed LDA\label{alg:Light-Partially-Collapsed}}
\end{algorithm}

\begin{algorithm}
\SetAlgoLined 
\SetKwData{Left}{left}
\SetKwData{This}{this}
\SetKwData{Up}{up} 

\SetKwFunction{Union}{Union}
\SetKwFunction{FindCompress}{FindCompress} 
\SetKwInOut{Input}{input}
\SetKwInOut{Output}{output}
\KwData{$\mathbf{A}$, $\mathbf{x}$, $\mathbf{z}$, $\sigma_a$, $\mathbf{n}^{(w)}$} \KwResult{$\mathbf{z}$,$\mathbf{n}^{(w)}$}

\BlankLine
\For{$d \leftarrow 1$ \KwTo $D$ in parallel}{      
   
   \For{$i \leftarrow 1$ \KwTo $N_d$}{
      // Remove position $i$ from the sufficient statistics

      $\mathbf{n}_{z_i}^{(d)}, \Delta \mathbf{n}_{z_i,x_i}^{(w)} \leftarrow -1$ \;

      // Word proposal draw

      $u_{i,1} \leftarrow$ RandomUniform()\;
      $z^* \leftarrow$ \texttt{LookUpAliasTable}($A_{x_i}$, $u_{i,1}$)\; 
      $u_{i,2} \leftarrow$ RandomUniform()\;
      \If{$\min\left\{ 1,\frac{\alpha+n_{d,z^{*}}^{-i}}{\alpha+n_{d,z_{i}}^{-i}}\right\} < u_{i,2}$}{
          $z_i \leftarrow z^*$\; 
      }
      // Document proposal draw

      $u_{i,3} \leftarrow$ RandomUniform()\;
      $u_{d,i,3} \leftarrow u_{i,3} \cdot(\alpha \cdot K + {N_{d,-i}})$\;
      \eIf{$u_{d,i,3} < \alpha \cdot K$}{
          // Normalize the random draw to [0,1]

          $u_{i,3} \leftarrow u_{d,i,3} / (\alpha \cdot K)$\;
          // Draw sample using $u_{i,3}$

          $z^* \leftarrow$ SampleDiscreteUniform(1, $K$, $u_{i,3}$) \; 
      }{
          // Normalize  to [0,1] $u_{i,3}$

          $u_{i,3} \leftarrow (u_{d,i,3} - (\alpha \cdot K)) / {N_{d,-i}}$\;
          $z^* \leftarrow$ ChooseRandomToken($\mathbf{z}_{- i}, u_{i,3}$) \; 
	  }
      $u_{i,4} \leftarrow$ RandomUniform()\;

      \If{$\min\left\{ 1,\frac{\phi_{k=z^{*},w_{i}=v}}{\phi_{k=z_{i},w_{i}=v}}\right\}  < u_{i,4}$}{
          $z_i \leftarrow z^*$ \;
      }

      // Add the topic indicator to the sufficient statistics

      $\mathbf{n}_{z_i}^{(d)}, \Delta \mathbf{n}_{z_i,x_i}^{(w)} \leftarrow +1$ \;
   }     
   $\mathbf{n}^{(w)} \leftarrow$ UpdateGlobalSufficientStatistics($\mathbf{n}^{(w)}$, $\Delta \mathbf{n}^{(w)}$) 
}

\caption{\texttt{SampleTopicIndicatorsLight()\label{alg:SampleTopicIndicatorsLight()}}}
\end{algorithm}

\begin{algorithm}
\SetAlgoLined 
\SetKwData{Left}{left} 
\SetKwData{This}{this} 
\SetKwData{Up}{up} 
\SetKwFunction{Union}{Union} 
\SetKwFunction{FindCompress}{FindCompress}
\KwData{vector of proportions $\mathbf{p}$, length of $\mathbf{p}$ $K$ } 
\KwResult{Alias table $A$}

\BlankLine
Initialize $L = U = \emptyset$ and $A = []$\;
\For{$i \leftarrow 1$ \KwTo $K$}{
	\eIf{p[i] $< 1 / K$}{
		$L \leftarrow L \cup (p[i], i)$\;
	}{
		$U \leftarrow U \cup (p[i], i)$\;
	}
}

\For{$i \leftarrow 1$ \KwTo $K$}{
	Get $(p_l, k_l)$ from $L$ and $(p_u, k_u)$ from $U$\;
	$A \leftarrow A \cup (p_l, k_l, k_u)$ \;
	$p_u \leftarrow p_u - (1/K - p_l)$ \;
	\eIf{$p_u < 1/K$}{
		$L \leftarrow L \cup (p_u, k_u)$\;
	}{
		$U \leftarrow U \cup (p_u, k_u)$\;
	}
}

\caption{\texttt{ConstructAliasTable()\label{alg:ConstructAliasTable()}}}
\end{algorithm}

\begin{algorithm}
\SetAlgoLined 
\SetKwData{Left}{left} 
\SetKwData{This}{this} 
\SetKwData{Up}{up} 
\SetKwFunction{Union}{Union} 
\SetKwFunction{FindCompress}{FindCompress}
\KwData{Alias table $A$, Alias table categories $K$, random uniform [0,1] $u$} 
\KwResult{Class $k$}

\BlankLine

$k \leftarrow \lfloor u \cdot K \rfloor + 1$ \;
// Renormalize $u$

$u \leftarrow (u \cdot K + 1 - k) / K$ \;
$(p, k_l, k_u) \leftarrow A[k]$ \;
\eIf{$u < p$}{
	return $k_l$ \;
}{
	return $k_u$ \;
}

\caption{\texttt{LookUpAliasTable()\label{alg:LookUpAliasTable()}}}
\end{algorithm}

\end{document}